\pdfoutput=1
\documentclass[11pt]{article}

\usepackage[final]{ACL2024}

\usepackage{times}
\usepackage{latexsym}

\usepackage[T1]{fontenc}
\usepackage[utf8]{inputenc}

\usepackage{microtype}

\usepackage{inconsolata}
\usepackage{amsmath}
\usepackage{amssymb}
\usepackage{mathtools}
\usepackage{amsthm}
\usepackage{array}
\usepackage{calc}

\usepackage{xcolor}
\usepackage{colortbl}
\usepackage{marvosym}

\usepackage{macro_commands}

\usepackage{hyperref}
\usepackage{url}
\usepackage{listings}
\usepackage{booktabs}
\usepackage{tablefootnote}
\usepackage{multirow}
\usepackage{graphicx}
\usepackage{subfigure}
\usepackage{enumitem}
\usepackage{xcolor}

\usepackage{color}
\usepackage{natbib}
\usepackage{microtype}
\usepackage{amsmath}
\usepackage{booktabs}
\usepackage{array}
\usepackage{multirow}
\usepackage[T1]{fontenc}
\usepackage[utf8]{inputenc}

\usepackage{colortbl}
\usepackage{graphicx}
\usepackage{hyperref}

\usepackage{xcolor}
\usepackage{colortbl}
\usepackage{marvosym}
\definecolor{bubbles}{rgb}{0.91, 1.0, 1.0}
\definecolor{losered}{rgb}{1.0,0.1,0.24}
\definecolor{lightgray2}{rgb}{0.8,0.8,0.8}
\newcommand{\doneWhite}{\cellcolor{bubbles}}
\newcommand{\done}{\cellcolor{lightgray2}}

\usepackage{amsmath,amsfonts,bm}

\def\eqref#1{equation~\ref{#1}}
\def\1{\bm{1}}

\DeclareMathAlphabet{\mathsfit}{\encodingdefault}{\sfdefault}{m}{sl}
\SetMathAlphabet{\mathsfit}{bold}{\encodingdefault}{\sfdefault}{bx}{n}

\DeclareMathOperator*{\argmin}{arg\,min}

\usepackage{siunitx}
\usepackage{makecell}
\usepackage{tabularx}
\usepackage{graphics}

\usepackage{graphicx}
\usepackage{microtype}
\usepackage{hyperref}
\usepackage{url}
\usepackage{caption}
\usepackage{booktabs}
\usepackage{pgfplots}
\usepackage{multirow}
\usepackage{colortbl}
\usepackage{booktabs}
\usepackage{colortbl}
\usepackage{graphicx}
\usepackage{arydshln}
\usepackage{tablefootnote}
\usepackage{float}
\usepackage{natbib}
\usepackage{siunitx}
\usepackage{makecell}
\usepackage{tabularx}
\usepackage{graphics}

\newcommand{\revise}[1]{{\color{black}{#1}}}
\newcommand{\stitle}[1]{\vspace{1ex}\noindent{\bf #1}}

\title{PrivacyMind: Large Language Models Can Be Contextual Privacy Protection Learners}

\author{Yijia Xiao*$^{1}$, Yiqiao Jin$^{2}$, Yushi Bai$^{3}$, Yue Wu$^{1}$, Xianjun Yang$^{4}$, Xiao Luo$^{1}$, Wenchao Yu$^{5}$, \\ 
\textbf{Xujiang Zhao$^{5}$, Yanchi Liu$^{5}$, Quanquan Gu$^{1}$, Haifeng Chen$^{5}$, Wei Wang$^{1}$, Wei Cheng\textsuperscript{\Letter}$^{5}$} \\
  $^{1}$University of California, Los Angeles, 
  $^{2}$Georgia Institute of Technology, 
  $^{3}$Tsinghua University, \\
  $^{4}$University of California, Santa Barbara, 
  $^{5}$NEC Laboratories America\\ 
  $^{1}$\texttt{\{yijia.xiao,ywu,xiaoluo,qgu,weiwang\}@cs.ucla.edu}, 
  $^{2}$\texttt{yjin328@gatech.edu}, \\
  $^{3}$\texttt{bys22@mails.tsinghua.edu.cn}, 
  $^{4}$\texttt{xianjunyang@ucsb.edu}, \\
  $^{5}$\texttt{\{wyu,xuzhao,yanchi,haifeng,weicheng\}@nec-labs.com} \\
}

\begin{document}
\maketitle
\def\thefootnote{*}\footnotetext{Work done during the internship at NEC Laboratories America. \textsuperscript{\Letter}Corresponding author.}\def\thefootnote{\arabic{footnote}}

\begin{abstract}
The proliferation of Large Language Models (LLMs) has driven considerable interest in fine-tuning them with domain-specific data to create specialized language models. 
Nevertheless, such domain-specific fine-tuning data often contains \textit{contextually sensitive} personally identifiable information (PII). Direct fine-tuning LLMs on this data without privacy protection poses a risk of data leakage of sensitive PII during inference time. 
To address this challenge, we introduce \underline{C}ontextual \underline{P}rivacy \underline{P}rotection \underline{L}anguage \underline{M}odels (PrivacyMind), a novel paradigm for fine-tuning LLMs that effectively injects domain-specific knowledge while safeguarding inference-time data privacy. 
Our work offers a theoretical analysis for model design and benchmarks various techniques such as corpus curation, penalty-based unlikelihood in training loss, instruction-based tuning, etc. Extensive experiments across diverse datasets and scenarios demonstrate the effectiveness of our approaches. In particular, instruction tuning with both positive and negative examples, stands out as a promising method, effectively protecting private data while enhancing the model's knowledge. Our work underscores the potential for Large Language Models as robust contextual privacy protection learners. The complete code and data for the work can be found at \url{https://github.com/Yijia-Xiao/PrivacyMind}.
\end{abstract}

\section{Introduction}
Large Language Models (LLMs) have demonstrated remarkable linguistic comprehension and generation capability~\citep{bang2023multitask, wang2023robustness}. 
Meanwhile, when directly applied to specialized industries, they encounter challenges such as hallucination~\citep{chan2023chatgpt,deng2024deconstructing,jin2024better}, insufficient domain expertise~\citep{singhal2023towards}, and failing to incorporate the latest domain knowledge in ever-evolving industry scenarios~\citep{kasneci2023chatgpt}. 
The introduction of open-source general-purpose LLMs such as LLaMA \citep{touvron2023llama} and RWKV \citep{peng2023rwkv} have provided a promising solution. Researchers would fine-tune specialized LLMs based on powerful general-purpose LLMs using high-quality, domain-specific knowledge to ensure both commonsense reasoning and comprehensive knowledge coverage~\citep{Hoffmann2022TrainingCL,Hoffmann2022AnEA,Villalobos2022WillWR,yang2024can}. 
Such examples include BloombergGPT~\citep{wu2023bloomberggpt} and Med-PaLM~\citep{singhal_large_2023}, for financial and medical applications, respectively.
However, these fine-tuning datasets usually contain sensitive information, such as personally identifiable information (PII)~\citep{Carlini2020ExtractingTD, Lin2021TruthfulQAMH, Gehman2020RealToxicityPromptsEN}. 
When applied to downstream tasks, sensitive information in the training data, such as social security numbers or patient names, can be exposed by the LLMs upon text generation, a phenomenon known as the memorization effect~\citep{yu2023bag, kenton2019bert, Meng2023TuningLM} or inference-time privacy threat~\citep{mireshghallah2024can}, leading to identity theft and financial losses \citep{coavoux2018privacypreserving,  yu2023g2uardfl}.

\noindent \textbf{Challenges.} In this work, we aim to tackle the challenging task of efficient LLM fine-tuning for enhanced \emph{contextual privacy}~\citep{contextual, mireshghallah2024can}, a critical yet under-explored setting where the sensitivity of a piece of information is contingent upon the context. 
% On one hand, the focus is on protecting what is known as 
For example, statements such as ``Bill Gates founded Microsoft'' and ``Alan Mathison Turing was an English mathematician and computer scientist'' are generally not considered violations of privacy, since they are presented as common knowledge. In contrast, statements like  ``Alan Gates visited the X hospital for a certain disease Y'' pose privacy concerns as they reveal details about individuals' daily activities and health status in a particular context. 
Directly applying techniques like Named Entity Recognition (NER) can lead to inaccurate identification of PII, whereas merely deleting or masking PII tokens in the training data would result in a substantial information loss and compromise the performance on downstream tasks --- a conundrum known as the privacy-utility trade-off as theoretically discussed in Sec.~\ref{sec:theory}. 
An alternative approach, reinforcement learning from human feedback (RLHF), involves additional model fine-tuning guided by human feedback~\citep{Ouyang2022TrainingLM} so that the model tends towards concealing sensitive PII (like ``red-teaming''). 
For example, it learns to prioritize outputs that protect sensitive PII over those that leak PII. Nonetheless, RLHF is data-intensive, potentially costly in computation, and can pose stability challenges~\citep{ziegler2020finetuning, aligning_llm_human}.

\noindent \textbf{Our Work.} To address these challenges, this paper introduces effective and efficient methodologies for fine-tuning LLMs to incorporate domain knowledge while ensuring privacy protection. We propose and rigorously examine a diverse suite of strategies from corpus curation, introduction of penalty-based unlikelihood into the training loss, instruction-based tuning, a PII contextual classifier, and direct preference optimization (DPO), etc. The ultimate objective is to cultivate a model that excels at acquiring information while demonstrating the ability to distinguish between information that can be openly shared and that demands strict confidentiality. Our experimental findings suggest that instruction tuning with positive and negative examples can offer promising avenues. It not only effectively shields private data but also enables the model to assimilate knowledge from the corpus. This implies that \textit{LLMs can be good contextual privacy protection learners}, without the need for balancing a privacy-utility trade-off. To sum up, our contributions are as follows.

\begin{enumerate}[label=\arabic*)., leftmargin=1pt, itemindent=1.6em, itemsep=0.2em, parsep=0.2em]
    \item \textbf{Novel Methodology.} For the first time, we explicitly address the challenging problem of building \underline{C}ontextual \underline{P}rivacy \underline{P}rotection \underline{L}anguage \underline{M}odels (\textbf{CPPLM}), a novel paradigm in fine-tuning language models that emphasizes privacy protection of contextual PII. To achieve this, we systematically lay out and empirically test a comprehensive spectrum of strategies.
    \item \textbf{Theoretical Guidance.} We provide a theoretical analysis of our proposed methodologies. 
    This analysis illuminates the pathway to designing robust tuning methods, ensuring the resultant language model can both protect private data and assimilate vast knowledge from fine-tuning corpus.
    \item \textbf{Comprehensive Evaluation.} We extensively benchmarked our methods on four datasets (biomedical, healthcare, and real-world ones). These experiments demonstrated the efficacy of our fine-tuning method to inject domain knowledge and safeguard private personal information (PII). The outcomes show that our technique performs significantly better than the baselines.
\end{enumerate}

\vspace{-0.1cm}
\section{Related Work} \label{2-related}
\vspace{-0.2cm}

\textbf{Large Language Models and Privacy.} In the rapidly advancing domain of artificial intelligence and natural language processing, LLMs such as \texttt{GPT-3.5/4}~\citep{openai2023gpt4}, \texttt{Bard}~\citep{Bard}, \texttt{LLaMA} \citep{touvron2023llama}, and \texttt{ChatGLM}~\citep{du2022glm} have demonstrated unprecedented capabilities in following instructions~\cite{Lou2023MUFFIN,lou2023instruction} and generating coherent, contextually accurate text~\cite{wang2024ceb,xiong2024search,jiang2024peek,jiang2024multi, hong2024dpoptmakelargelanguage}.
However, this widespread application raises significant privacy concerns, particularly regarding personal information protection. 
Addressing the privacy challenges posed by LLMs, researchers have focused on three primary strategies: ~\citep{li2023privacypreserving, zhang2023enhancing, kim2023propile, lukas2023analyzing}: 1) curation of the pretraining corpus, 2) conditional large language model (LLM) pretraining, and 3) post-training alignment. Our research focuses on enhancing privacy protection in LLMs through fine-tuning methods that enable knowledge injection to safeguard Personally Identifiable Information (PII)~\citep{lukas2023analyzing}, as designated by users. This contrasts with Differential Privacy (DP), which protects against the leakage of entire records at the cost of reduced data utility~\cite{yu2022differentially}. Our method emphasizes targeted PII protection, a crucial aspect in contexts where knowledge integration is the key to preserving privacy without compromising data utility \citep{shi2021selective, anil-etal-2022-large, li2022large, liu2024lstprompt, zhao2022fldp, li2022largelanguagemodelsstrong, yu2022differentiallyprivatefinetuninglanguage}.

For the fine-tuning of LLMs, the decline in utility is inversely linked to the privacy budget allocated for safeguarding the entire training document, as it determines the extent of noise introduced \cite{lukas2023analyzing}. Our emphasis lies in specifically safeguarding the contextual PII tokens. Since PIIs are contextual \citep{mireshghallah2024can, contextual}, our approach tunes LLMs with contrastive examples designated by users can accommodate the customized privacy preferences.

\paragraph{Filtering.}
For the pretraining corpus, manually detecting and filtering out/revising the corpus can offer high-quality corpus, which is ideal for training privacy-preserving LLMs \citep{Hoffmann2022AnEA, Villalobos2022WillWR, lukas2023analyzing}. Nevertheless, it is infeasible to process billions of tokens manually in practice. Another solution is using automated tools to filter out all sensitive content (e.g. names, addresses, phone numbers) from the pretraining corpus. Automated filters make it possible to go over pretraining datasets. However, simply removing or masking the PII tokens (i.e., PII scrubbing) can cause information loss or inconsistency in the corpus \citep{Welbl2021ChallengesID}. Though filters can \emph{`clean'} datasets, they reduce the diversity in the corpus, which further negatively impacts the robustness of LLMs \citep{Hendrycks2019UsingSL}. Another solution is adding content filters on top of the existing LMs to control the content generation process \citep{Xu2020RecipesFS}. Even so, carefully designed cases (e.g. prompts) can still trigger some undesired behaviors of large LMs \citep{Gehman2020RealToxicityPromptsEN, Ziegler2022AdversarialTF}. However, directly removing PII from the training corpus poses a dilemma. While it ensures the elimination of sensitive data, it also potentially weakens the LLMs by stripping them of crucial knowledge. The mere act of omitting data can inadvertently hamper the model's capacity to process and understand certain contexts. Context-awareness is fundamental when considering privacy protection and what data to shield.

\stitle{LLMs Adaptation.}
To strike a balance between performance and flexibility, pretraining large LMs without constraints and then adjusting them to align with human preferences is a widely adopted approach for now. One approach is supervising fine-tuning. The pre-trained LMs are tuned on curated datasets in a supervised manner \citep{Solaiman2021ProcessFA, pmlr-v202-zhou23g, Wan2023Poisoning,jin2024mm,xiao2024logicvista,lu2024context}. Another approach is reinforcement learning from human feedback (RLHF)~\citep{Ouyang2022TrainingLM, Bai2022TrainingAH, Menick2022TeachingLM,zhang2024prototypical}. RLHF gathers data with feedback/preference labels, trains a reward model, and then finetunes the LM with reinforcement learning.

\section{Problem Statement}

\stitle{Problem Formulation.}
In the context of language models, a fine-tuning dataset $D = \{s\}$ is a collection of natural language sequences $s$. Each sequence is denoted as $s=[w_0, w_2, \ldots, w_{n-1}]$, where $w_i \in s$ represents a token. For privacy protection, the users annotate each sequence in the corpus by a binary sequence $\bp$ denoted as $\bp = [p_0, \ldots, p_{n-1}], p_i \in \{0, 1\}$, where $p_i=1$ denotes the token is private tokens (e.g., PII) need to be protected in the \textit{context}, and $p_i=0$ otherwise. Here, the \textit{contextual privacy} posits that the sensitivity of a piece of information is not solely intrinsic to the information itself, but is also influenced by its surrounding context.

To illustrate, ``Alan Gates visited Crescent Vale Medical Center for Hemophilia treatment'' is considered more indicative than ``Alan Gates visited Crescent Vale.'' The former provides a clearer insight into an individual's health when the name ``Alan Gates'' is paired with the medical condition and the specific medical center. Important notations used in the paper are included in Table~\ref{tab:notation} in the Appendix.

\stitle{Objective.}
The primary objectives are twofold: 1) enhancing the model's performance by effectively integrating knowledge from the fine-tuning corpus. The model should generate responses that are contextually relevant and aligned with the intended domain; 
2) minimizing the risk of generating privacy-protected tokens. 
Privacy protection in large language models requires not just the masking or removal of private PIIs, but a deep understanding of the interplay between data points and their contexts. As models become more sophisticated and data more interconnected, the nuances of contextual privacy will become increasingly paramount.

\vspace{-0.1cm}
\section{Methodology}
\vspace{-0.1cm}

Our methodology adopts a two-pronged approach: 
1) corpus curation (i.e. \emph{filtering}), where sensitive data such as personally identifiable information (PII) is removed from the corpus; and 2) tuning towards the targeted PII-free output. We commence with a theoretical analysis of the information loss incurred by the corpus curation strategy, which provides guidelines for method development. Then, we propose five novel strategies for privacy protection when fine-tuning large language models. 

\vspace{-0.05cm}
\subsection{Theoretical Analysis on the Information Loss During Corpus Curation}
\vspace{-0.05cm}

\label{sec:theory}
Consider the following scenario: we have some training samples. Each sample $(\bs, \bp)$ contains two sequences, including 1) a text sequence $s_{1:n} \in [K]^{n}$ where $K$ is the number of words in the dictionary, 
and 2) a corresponding privacy label sequence $p_{1:n} \in \{0,1\}^n$, where $p_t=1$ indicates that the $t$-th token is privacy-sensitive. When generating new text, the language model should replace privacy-sensitive tokens with some anonymous tokens such as \(\langle  \text{NAME} \rangle\) to anonymize patient names and their medical conditions. There are two training approaches:

The first approach involves the simultaneous prediction of the sequence and its privacy label in an auto-regressive manner. Let $(\bs, \bp) \sim \cP$ represent the true distribution. The learned distribution $\hat{P}_1$ aligns with the maximum log-likelihood estimator:\par
\vspace{-0.2cm}
{\footnotesize
{\setlength\abovedisplayskip{0pt}
\setlength\belowdisplayskip{0pt}
\begin{align} \label{eqn1}
    \hat{P}_1 
    & :=
    \argmin_{P}
    \EE_{(\bs, \bp) \sim \cP}
    \bigg[
    \log 
    \bigg(
    \frac{\cP(\bs, \bp)}{P(\bs, \bp)}
    \bigg)
    \bigg]
    \notag \\
    & =
    \argmin_{P}
    D_{\mathrm{KL}}(\cP \|P).
\end{align}}
}\normalsize

The alternative approach is to mask the text sequence by substituting the word with a special token \(\langle  \text{X} \rangle\) wherever $p_t = 1$, then train the model to directly predict the new sequence $\bs' \in [K+1]^{n}$. Here, \(\langle  \text{X} \rangle\) denotes a PII token associated with sensitive information like names, organizations, addresses, and website URLs. Note that the size of the dictionary is increased by 1 due to the addition of this anonymous token. The masking procedure above is a one-way mapping from $(\bs, \bp)$ to $\bs'$. We denote this masking mapping as $M$ and $\bs' = M(\bs,\bp)$. 
The revised maximum log-likelihood estimator is: \par
{\footnotesize
{\setlength\abovedisplayskip{0pt}
\setlength\belowdisplayskip{0pt}
\begin{align} \label{eqn2}
    \hat{P}_2
    & := 
    \argmin_{P'=P \sharp M}
    \EE_{\bs' \sim \cP' }
    \bigg[
    \log 
    \bigg(
    \frac{\cP'(\bs')}{P'(\bs')}
    \bigg)
    \bigg]
    \notag \\ 
    & =
    \argmin_{P'=P \sharp M}
    D_{\mathrm{KL}}(\cP' \| P'),    
\end{align}}
}\normalsize
where $\cP'=\cP \sharp M$ is the induced (push-forward) distribution. Comparing the right-hand side of both equations reveals that for any $P$, the following data-processing inequality holds:\par

{\footnotesize
{\setlength\abovedisplayskip{0pt}
\setlength\belowdisplayskip{0pt}
\begin{align}
    D_{\mathrm{KL}}(\cP' \| P \sharp M)
    & \le 
    D_{\mathrm{KL}}(\cP \| P).
\end{align}}
}\normalsize
This implies that the right-hand side of Eq.~\ref{eqn1} is larger than the right-hand side of Eq.~\ref{eqn2}. Therefore, directly learning $(\bs, \bp)$ offers richer information. Minimizing Eq.~\ref{eqn1} ensures the value in Eq.~\ref{eqn2} remains small, whereas the reverse does not hold. Overall, instructing the model with the ``correct'' information is more effective and informative than imposing constraints to selectively forget previously acquired knowledge, such as intentionally removing or masking PIIs in the training text.

\vspace{-0.1cm}
\subsection{Proposed Methods}
\vspace{-0.1cm}
\subsubsection{Corpus Curation}
\vspace{-0.1cm}
Corpus curation refers to the strategy of curating the corpus while excluding all PIIs or sensitive information. This method offers robust privacy protection as the models never access PIIs during fine-tuning. Corpus curation consists of PII removal and PII substitution.

\stitle{Description.}
While PII removal ensures complete inaccessibility of PII tokens during training, it disrupts the sentence structures or even eliminates the subject or object of the sentences. Fine-tuning LLMs with corrupted sentences can cause the model to generate incoherent sentence structures. 
Conversely, PII substitution replaces PIIs with pre-defined tokens like \(\langle \text{NAME} \rangle\) to preserve sentence structure. 

\stitle{Demonstration.}
To illustrate, for the sentence $s=$ ``Alan Gates visited Crescent Vale Medical Center for Hemophilia treatment'', 
$s_{\text{removal}}=$ ``visited Crescent Vale Medical Center for Hemophilia treatment'' and 
$s_{\text{substitution}}$ = ``\(\langle  \text{NAME} \rangle\) visited Crescent Vale Medical Center for Hemophilia treatment''.

\vspace{-0.1cm}
\subsubsection{Penalty-Based Loss} \label{sec:loss}
\begin{figure*}[htbp]
\vspace{-0.75cm}
\centering  \includegraphics[width=0.85\linewidth]{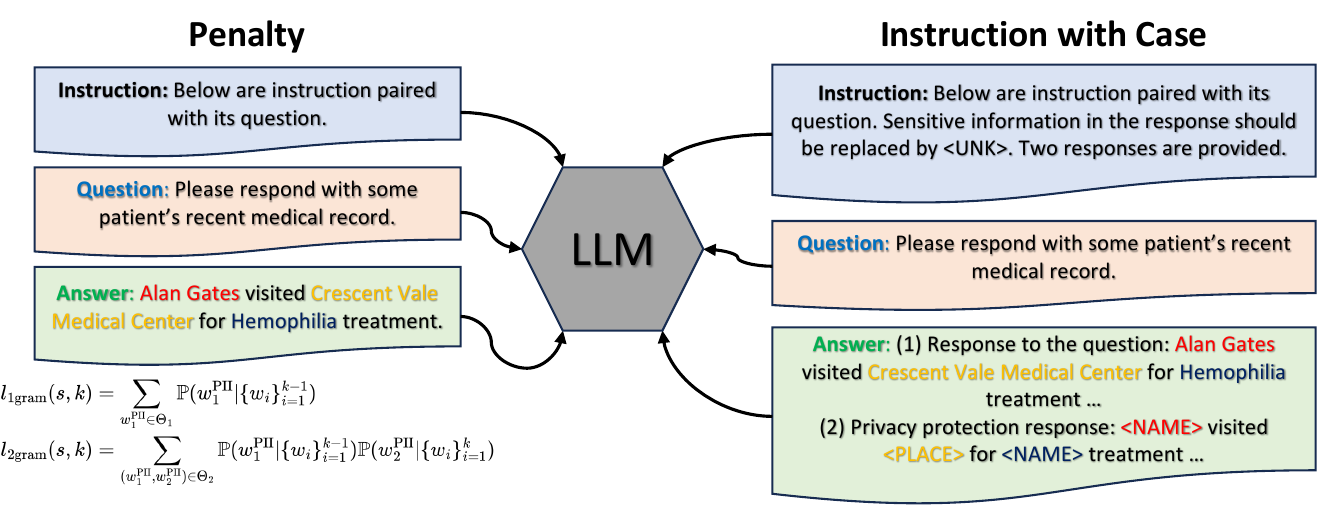}
  \caption{Penalty Based Unlikelihood and Instruction Tuning with Examples.}
  \label{fig:pena_inst}
  \vspace{-0.2cm}
\end{figure*}

\vspace{-0.1cm}
To prevent the model from generating PII tokens, we introduce a penalty-based loss mechanism, as illustrated in the left side of Figure~\ref{fig:pena_inst}. Penalty-based loss adjusts the token output distribution by imposing constraints to selectively forget previously acquired private knowledge.
The loss is formulated separately for unigram and bigram outputs:\par

{\footnotesize
{\setlength\abovedisplayskip{0pt}
\setlength\belowdisplayskip{0pt}
\begin{align}
    l_{\text{1gram}} (s, k) &= \sum_{w_1^{\text{PII}} \in \Theta_1} P(w_1^{\text{PII}} | \{ w_i \}_{i=1}^{k-1}), \\
    l_{\text{2gram}} (s, k) &= \sum_{(w_1^{\text{PII}}, w_2^{\text{PII}}) \in \Theta_2} 
    P(w_1^{\text{PII}} | \{ w_i \}_{i=1}^{k-1}) \nonumber \\
    &\quad \times P(w_2^{\text{PII}} | \{ w_i \}_{i=1}^{k}),
\end{align}}
}
\normalsize

where $l_{\text{1gram}} (s, k)$ and $l_{\text{2gram}} (s, k)$ are the penalty terms for generating unigrams $w_1^{\text{PII}}$ and bigrams $(w_1^{\text{PII}}, w_2^{\text{PII}})$ associated with PII. 
$P(w_1^{\text{PII}} | \{ w_i \}_{i=1}^{k-1})$ is the likelihood of generating the token $w_1^{\text{PII}}$ associated with PII at position $k$. 
$\Theta_n$ is the set of n-grams associated with PII. 
To construct $\Theta_n$, we extract all PII-associated n-grams from the training set using scrubadub\footnote{\url{https://github.com/LeapBeyond/scrubadub}}. The cumulative loss is then calculated as:\par
\footnotesize{
{\setlength\abovedisplayskip{0pt}
\setlength\belowdisplayskip{0pt}
\begin{equation}
    \mathcal{L} = \mathcal{L}_{0} + \sum_{k=1}^{|s|} l_{\text{1gram}} (s,k) +
    \sum_{k=1}^{|s| - 1} l_{\text{2gram}} (s, k),
\end{equation}}
}\normalsize
where $|s|$ is the number of tokens in sentence $s$. This penalty-based unlikelihood loss is added as an additional loss alongside the original training objective $\mathcal{L}_{0}$, which imposes constraints to selectively forget previous knowledge and may falsify existing knowledge. Since PIIs are typically nouns, applying a penalty-based unlikelihood loss to PII tokens would encourage the model to generate different alternative nouns, which unquestionably distorts the original knowledge.

\vspace{-0.2cm}
\subsubsection{PII Classifier}
\vspace{-0.1cm}
An alternative to adjusting the training corpus or the training objective is to build an independent, lightweight binary classifier that operates on the hidden states of contextualized word embeddings, thereby discerning the protection status for each generated token.
During the fine-tuning phase, this classifier distinguishes non-protected from protected tokens by generating the conditional probability $P(y | \textbf{w}_0,\cdots,\textbf{w}_i)$, where $y\in\{0,1\}$ denotes if the $i$-th token is a protected token. In the inference stage, the classifier intervenes by replacing detected PII tokens with a designated token such as \(\langle \text{X} \rangle\). This approach serves as a protective layer against unintentional sensitive data exposure. Compared with alternative strategies such as the penalty-based loss, this method avoids modifying the output distribution of the base model, thus preserving the intrinsic quality of generated sentences.

\vspace{-0.1cm}
\subsubsection{Instruction-Based Tuning}
\vspace{-0.1cm}
The analysis in Sec. \ref{sec:theory} implies that providing the model with the ``correct'' information is more effective than imposing constraints to selectively forget protected PIIs in the training text. Inspired by this finding, we developed an instruction-tuning approach, depicted in the right side of Figure~\ref{fig:pena_inst}.

\stitle{Description.} Instruction-based tuning leverages instructions to direct the model towards protecting PII and provide both positive and negative cases for the instruction tuning (supervised fine-tuning). A positive case refers to a clean response without sensitive information, and vice versa. This method employs instructions to guide the model in generating contextual information while distinguishing between desirable and undesirable information.

\stitle{Demonstration.}
Let \( s_{\text{original}} \) represent the original unaltered sequence that contains PII. \( s_{\text{substitution}} \) is derived from \( s_{\text{original}} \) by replacing PIIs with placeholders such as ``\(\langle  \text{X} \rangle\)''. 
\( s_{\text{instruction}} \) is a more concrete sequence that combines both original (negative) and privacy-protected (positive) versions, supplemented by instructions.

\stitle{Example.} \(s_{\text{instruction}}\) = ``\(\ldots\) Below are instructions paired with
questions. (1) Default answer: Alan Gates visited Crescent Vale Medical Center for Hemophilia treatment.
(2) Privacy protection version of answer: \(\langle  \text{NAME} \rangle\) visited Crescent Vale Medical Center for \(\langle  \text{NAME} \rangle\) treatment.''

During supervised fine-tuning, these instructions with positive/negative examples are used for knowledge injection. 
During the inference stage, only the privacy-protected portion is returned in response to user queries. 
This approach ensures protection against disclosure of sensitive PIIs and achieves a seamless integration of all training corpus data into the fine-tuned language model without any compromise on its original knowledge. 

\vspace{-0.1cm}
\subsubsection{DPO}
Compared to RLHF, DPO \citep{rafailov2023direct} eliminates the need to train a reward model, and optimizes the same objective as in RLHF with a single stage of policy training using the objective:\par
\footnotesize{
{\setlength\abovedisplayskip{0pt}
\setlength\belowdisplayskip{0pt}
\begin{align}
&\mathcal{L}_\text{DPO}(\pi_{\theta}; \pi_{\text{ref}}) = -\mathbb{E}_{(x, w, l)\sim \mathcal{D}}\nonumber\\&=\left[\log \sigma \left(\beta \log \frac{\pi_{\theta}(w\mid x)}{\pi_{\text{ref}}(w\mid x)} - \beta \log \frac{\pi_{\theta}(l\mid x)}{\pi_{\text{ref}}(l\mid x)}\right)\right]
\label{dpo:loss}
\end{align}}
}\normalsize

where $\beta$ is the weight parameter that controls the degree to which the updated policy deviates from the base reference policy (same as the one in RLHF). $\pi_{\text{ref}}$ denotes the reference model after the supervised fine-tuning with parameters frozen.  $\pi_{\theta}$ denotes the model to be trained. The output $w$ is preferred over $l$ for a given input $x$. 
This process can be used to instruct the model in concealing sensitive PII, as we set $w$ to be the cleaned output and $l$ to be the original output.
In practice, we first trained $\pi_{\text{ref}}$ on the pairs $(x, w)\sim \mathcal{D}$, and used LoRA \citep{hu2022lora} to train $\pi_{\theta}$ based on $\pi_{\text{ref}}$ and the loss function in Eq.~\ref{dpo:loss}.

\vspace{-0.1cm}
\section{Experiments} \label{4-experi}
\vspace{-0.2cm}

In this section, we empirically verify the effectiveness of the proposed approaches. Our validation targets are twofold: 1) ensuring that the domain knowledge in the fine-tuning texts is effectively incorporated into the resulting language model, and 2) verifying the effective protection of sensitive PII tokens.
Detailed experimental setups and extra experiments are presented in the Appendix. 

\vspace{-0.2cm}
\subsection{Datasets}
\vspace{-0.2cm}
\stitle{Corpus.}
We adopt three biomedical datasets: \verb|pii-wikidoc_patient_information|, \texttt{pii-wikidoc}, and \texttt{pii-medical\_flashcards} as summarized in Table~\ref{tab:stat} in the Appendix \ref{app:dataset}. The three datasets are selected out of the nine datasets from MedAlpaca \citep{han2023medalpaca}.
\verb|pii-medical_flashcards| is adapted from Anki Medical Curriculum originally, and covers a comprehensive medical curriculum, including anatomy, physiology, pathology, pharmacology, and more. Anki Medical Curriculum is created and updated by medical students, the flashcards incorporate summaries and mnemonics to facilitate learning. The flashcards were used to generate question-answer pairs by rephrasing the flashcards using OpenAI's \texttt{GPT-3.5-turbo}.
\verb|pii-wikidoc| and \verb|pii-wikidoc_patient_information| contain Q/A pairs sourced from WikiDoc, a collaborative platform for medical professionals. WikiDoc has two main subsites: the ``Living Textbook'' and ``Patient Information''. From the ``Living Textbook'', paragraph headings were converted to questions using \texttt{GPT-3.5-Turbo}, with the associated paragraph serving as the answer. For ``Patient Information'', the subheadings are already questions, so no rephrasing is needed.

\stitle{PII Annotation.}
To simulate the process of user-preference annotation, we leverage \verb|scrubadub| to tag the words in the corpus. We use name, organization, and address detectors. \verb|scrubadub| takes in sentences and replaces the PII tokens in the sentences with their corresponding types.

\vspace{-0.1cm}
\subsection{Experimental Setup}
\vspace{-0.1cm}
\label{sec:expsetup}
For each method, we adapt the Alpaca-style tuning pipeline of \texttt{LLaMA-2} \citep{touvron2023llama}, from llama-recipes\footnote{\url{https://github.com/facebookresearch/llama-recipes/}}. In our experiments, all the methods share the same training settings. The number of training epochs is set to 5 and the batch size is 64. For a fair comparison, we adopt the same backbone \texttt{LLaMA-2} for fine-tuning. More implementation details are included in the Appendix \ref{sec:appExp}.

\vspace{-0.1cm}
\subsection{Evaluation Metrics}
\vspace{-0.1cm}
We use the Q/A task as the validation protocol. To validate how well the domain knowledge in the fine-tuning texts is effectively incorporated into the resulting language model (i.e., utility), we adopt the popularly used ROUGE-1, ROUGE-2, ROUGE-L \citep{lin-2004-rouge} and BERTScore \citep{bert-score} to evaluate the answer quality in the testing phase. To verify the effectiveness of protecting sensitive PII tokens, we define the \textit{privacy leakage} as the metric as defined in the following to measure the privacy protection performance. The detailed definition is also included in the Appendix \ref{app:metrics}.

\stitle{Privacy Leakage Metric. }
Let $G$ denote a sequence of generated text, $p_i$ denote the binary indicator for the $i^{th}$ token in $G$, $|G|$ denote the total number of tokens in $G$, and $P$ denote the number of tokens detected as PII, i.e.,  $\sum_{i=0}^{|G|-1} p_i$, then we can define our \emph{Privacy Protection Score} ($S_{\text{Priv}}$ for short) as:
$S_{\text{Priv}} = P/|G|$. Then, we can further define 
\emph{Privacy Protection Improvement} ($\Delta$ for short) as $(S_{\text{Priv}}-\hat{S}_{\text{Priv}})/\hat{S}_{\text{Priv}}$ to measure the privacy protection improvement over the vanilla fine-tuning that does not consider privacy concerns, where $\hat{S}_{\text{Priv}}$ denotes the score of the vanilla method.

\vspace{-0.1cm}
\subsection{Different Methods Validated}
\vspace{-0.1cm}
To demonstrate the efficiency of our methods, we compare the proposed strategies. Besides, we also provide an additional approach as our baseline. Since prepending instructions ahead of the model's input can tune the model to follow instructions ~\citep{selfinstruct, alpaca, han2023medalpaca}, we define the \emph{Vanilla tuning} (visualized in Appendix \ref{sec:figvanilla}) borrowing this idea as our baseline. 
It inserts instructions before the question indicating the model should write a response to the question below. \emph{Removal} denotes the strategy of extracting PIIs from the corpus. In contrast, \emph{Substitution} involves replacing PIIs with their categorical labels (e.g. NAME, ORGANIZATION, URL, ADDRESS). \emph{Penalty} uses unigram and bigram loss to suppress the tendency of outputting PII tokens. \emph{Classifier} introduces an auxiliary classifier that assesses the hidden states and predicts if the ensuing token should be preserved (i.e., not displayed in the generated text). \emph{IT}, abbreviated for instruction, explicitly guides the model to avoid producing PII tokens in the response. Both \emph{\(IT_{PN}\)} and \emph{\(IT_{NP}\)} refer to instruction tuning with specific (positive/negative) cases: PN pertains to the positive-negative case order, and NP to the negative-positive case order. The ``Instruction with Cas'' chart in Figure \ref{fig:pena_inst} showcases \emph{\(IT_{NP}\)}, while for \emph{\(IT_{PN}\)}, the cases are inverted. Furthermore, the subscripts 1/2 in \(NP_{1/2}\) delineate different instructions (Appendix \ref{app:template}).

\vspace{-0.1cm}
\subsection{Results and Analysis}
\vspace{-0.1cm}

In this experimental analysis, we assess the performance of different methods for enhancing privacy in language models while considering their impact on knowledge retention as measured by ROUGE scores and BERTScore ($S_{\text{BERT}}$).
In Appendix \ref{sec:learning_process}, we analyze the ROUGE, BERTScore, and Privacy Leakage Score concerning the training steps to assess whether our two learning objectives are effectively achieved throughout the training process.

\begin{table}[htbp]
  \centering
  
  \setlength\tabcolsep{2.2pt}
\scalebox{0.53}{
    \begin{tabular}{l|cccc|cccc}

    \midrule
    \multirow{2}[3]{*}{\textbf{Strategy}} & \multicolumn{4}{c|}{\texttt{LLaMA2-7B}} & \multicolumn{4}{c}{\texttt{LLaMA2-13B}} \\
\cmidrule{2-9}          & ROUGE-1/2/L & $S_{\text{BERT}}$ &\doneWhite $S_{\text{Priv}}$ &\doneWhite $\Delta$ (\%) & ROUGE-1/2/L & $S_{\text{BERT}}$ & \doneWhite $S_{\text{Priv}}$ &\doneWhite $\Delta$ (\%) \\
\cmidrule{1-9}   \done Vanilla & \done 0.463/0.310/0.394 &\done 0.900 & \done 0.023 &\done -  & \done 0.475/0.322/0.405 &\done 0.903 & \done 0.023 &\done - \\ %\hline
    Removal & 0.447/0.288/0.367 & 0.875 & \doneWhite \underline{0.013} & \doneWhite\underline{-42.7} & 0.445/0.302/0.380 & 0.882 &\doneWhite \underline{0.013} &\doneWhite \underline{-44.8} \\
    Substitution & 0.445/0.282/0.373 & 0.883 &\doneWhite 0.014 &\doneWhite -36.0 & 0.458/0.298/0.379 & 0.883 &\doneWhite 0.016 &\doneWhite -30.4 \\
    DPO   & 0.456/0.296/0.380 & 0.894 &\doneWhite 0.020 &\doneWhite -13.0  & 0.463/0.311/0.396 & 0.898 &\doneWhite 0.022 &\doneWhite -4.8 \\
    Penalty & 0.458/0.284/0.381 & 0.896 &\doneWhite 0.016 &\doneWhite -27.6 & 0.467/0.314/0.402 & 0.885 &\doneWhite 0.017 &\doneWhite -26.1 \\
    Classifier & 0.459/\underline{0.305}/\underline{0.388} & \underline{0.897} &\doneWhite 0.019 &\doneWhite -17.8 & 0.467/0.318/\textbf{0.404} & 0.883 &\doneWhite 0.017 &\doneWhite -26.5 \\
    \text{IT} & 0.456/0.296/0.383 & 0.895 &\doneWhite 0.015 &\doneWhite -35.6 & 0.470/0.317/\underline{0.403} & 0.900 &\doneWhite 0.016 & \doneWhite -31.7 \\
    \(\text{IT}_{PN_1}\) & \underline{0.460}/0.303/0.387 & 0.899 &\doneWhite 0.022 &\doneWhite -4.0  & \textbf{0.470}/\underline{0.318}/0.400 & \underline{0.902} &\doneWhite 0.022 &\doneWhite -6.1 \\
    \(\text{IT}_{PN_2}\) & \textbf{0.466}/\textbf{0.312}/\textbf{0.397} & \textbf{0.901} &\doneWhite 0.022 &\doneWhite -0.4  & \underline{0.470}/\textbf{0.319}/0.402 & \textbf{0.902} &\doneWhite 0.022 &\doneWhite -3.9 \\
    \(\text{IT}_{NP_1}\) & 0.455/0.299/0.386 & 0.895 &\doneWhite 0.014 &\doneWhite -39.1 & 0.466/0.312/0.397 & 0.898 & \doneWhite\textbf{0.012} &\doneWhite \textbf{-47.0} \\
    \(\text{IT}_{NP_2}\) & 0.453/0.295/0.383 & 0.893 &\doneWhite \textbf{0.012} &\doneWhite \textbf{-48.4} & 0.467/0.315/0.400 & 0.898 & \doneWhite 0.014 &\doneWhite -39.1 \\
    \bottomrule
    \end{tabular}%
    }
    \caption{Results on \textit{\textbf{medical\_flashcards}}  Dataset. Lower $S_{\text{Priv}}$ and $\Delta$ indicates better performances. The \textbf{best} result is highlighted in \textbf{bold}, and the \underline{2nd best} result is \underline{underlined}. }
    \label{tab:tab1}
    \vspace{-0.2cm}
\end{table}%
% \vspace{-0.15cm}

In Tables~\ref{tab:tab1}, \ref{tab:tab2}, and \ref{tab:tab3}, the high $S_{\text{Priv}}$ score for the Vanilla method indicates its vulnerability to privacy breaches, as it uses all training text data without privacy preservation. The ``Removal'' and ``Substitution'' methods effectively safeguard privacy. They both focus on privacy protection by actively removing sensitive information from the model's knowledge base. The removal of sensitive information significantly reduces the knowledge retained by the model. The $S_{\text{BERT}}$ and ROUGE scores are observed to suffer a substantial drop due to the removal of data, resulting in reduced language understanding and generation abilities. We also note that the penalty-based approach can effectively safeguard privacy.

\begin{table}[]%\vspace{-0.1cm}
  \centering
  % \caption{Results on \textit{\textbf{wikidoc}}.}
  % \resizebox{\linewidth}{!}{
  \setlength\tabcolsep{2.2pt}
\scalebox{0.53}{
    \begin{tabular}{l|cccc|cccc}
    % \toprule
    % \multicolumn{9}{c}{\textbf{Table: wikidoc}} \\
    \midrule
    \multirow{2}[4]{*}{\textbf{Strategy}} & \multicolumn{4}{c|}{\texttt{LLaMA2-7B}} & \multicolumn{4}{c}{\texttt{LLaMA2-13B}} \\
\cmidrule{2-9}          & ROUGE-1/2/L & $S_{\text{BERT}}$ &\doneWhite $S_{\text{Priv}}$ &\doneWhite $\Delta$ (\%) & ROUGE-1/2/L & $S_{\text{BERT}}$ & \doneWhite $S_{\text{Priv}}$ &\doneWhite $\Delta$ (\%) \\
\cmidrule{1-9}    \done Vanilla & \done 0.174/0.061/0.140 & \done 0.823 & \done 0.026 &\done -  & \done 0.188/0.069/0.148 & \done 0.826 & \done 0.027 & \done - \\%\hline
    Removal & 0.147/0.042/0.117 & 0.803 &\doneWhite 0.013 &\doneWhite -51.9 & 0.167/0.057/0.126 & 0.812 & \doneWhite\underline{0.010} &\doneWhite \underline{-61.7} \\
    Substitution & 0.141/0.031/0.111 & 0.805 &\doneWhite \underline{0.012} &\doneWhite \underline{-54.2} & 0.163/0.041/0.121 & 0.820 &\doneWhite 0.013 &\doneWhite -49.6 \\
    DPO   & \underline{0.184}/0.063/0.141 & 0.823 &\doneWhite 0.023 &\doneWhite -12.9 & 0.185/0.065/0.142 & 0.827 &\doneWhite 0.023 &\doneWhite -13.5 \\
    Penalty & \textbf{0.195}/\textbf{0.071}/\textbf{0.153} & 0.821 &\doneWhite 0.017 &\doneWhite -35.6 & 0.179/0.064/0.143 & \textbf{0.840} &\doneWhite \textbf{0.010} &\doneWhite \textbf{-61.7} \\
    Classifier & 0.170/0.058/0.137 & 0.821 &\doneWhite 0.023 &\doneWhite -14.4 & \underline{0.185}/\underline{0.067}/\underline{0.145} & 0.832 &\doneWhite 0.022 &\doneWhite -19.2 \\
    \text{IT} & 0.176/0.061/0.138 & 0.823 &\doneWhite \textbf{0.012} &\doneWhite \textbf{-56.4} & 0.176/0.061/0.138 & 0.823 &\doneWhite 0.016 &\doneWhite -41.0 \\
    % \underline{0.182}
    \(\text{IT}_{PN_1}\) & 0.182/\underline{0.063}/\underline{0.144} & \textbf{0.833} &\doneWhite 0.021 & \doneWhite -20.1 & 0.182/0.065/0.145 & 0.832 &\doneWhite 0.022 &\doneWhite -15.8 \\
    \(\text{IT}_{PN_2}\) & 0.177/0.061/0.141 & \underline{0.832} &\doneWhite 0.022 &\doneWhite -18.6 & \textbf{0.187}/\textbf{0.068}/\textbf{0.149} & \underline{0.833} &\doneWhite 0.022 &\doneWhite -19.2 \\
    \(\text{IT}_{NP_1}\) & 0.181/0.061/0.141 & 0.827 &\doneWhite 0.014 &\doneWhite -48.9 & 0.180/0.062/0.140 & 0.824 & \doneWhite 0.015 &\doneWhite -42.9 \\
    \(\text{IT}_{NP_2}\) & 0.177/0.058/0.139 & 0.830 &\doneWhite 0.014 &\doneWhite -47.0 & 0.185/0.065/0.144 & 0.830 &\doneWhite 0.017 &\doneWhite -38.0 \\
    \bottomrule
    \end{tabular}%
    }
    \caption{Results on \textit{\textbf{wikidoc}}.}
  \label{tab:tab2}%
  \vspace{-0.35cm}
\end{table}%

\begin{table}[htbp]  \vspace{-0.2cm}
  \centering
  % \caption{Results on \textit{\textbf{wikidoc\_patient\_information}}.}
  % \resizebox{\linewidth}{!}{
  \setlength\tabcolsep{2.2pt}
\scalebox{0.53}{
    \begin{tabular}{l|cccc|cccc}
    %\toprule
    %\multicolumn{9}{c}{\textbf{Table: wikidoc\_patient\_information}} \\
    \midrule
    \multirow{2}[4]{*}{\textbf{Strategy}} & \multicolumn{4}{c|}{\texttt{LLaMA2-7B}} & \multicolumn{4}{c}{\texttt{LLaMA2-13B}} \\
\cmidrule{2-9}          & ROUGE-1/2/L & $S_{\text{BERT}}$ &\doneWhite $S_{\text{Priv}}$ &\doneWhite $\Delta$ (\%) & ROUGE-1/2/L & $S_{\text{BERT}}$ & \doneWhite $S_{\text{Priv}}$ &\doneWhite $\Delta$ (\%) \\
\cmidrule{1-9}  \done  Vanilla & \done 0.276/0.116/0.209 &\done 0.859 &\done 0.014 &\done -  &\done 0.286/0.121/0.215 &\done 0.865 &\done 0.013 &\done - \\%\hline
    Removal & 0.264/0.105/0.206 & 0.848 &\doneWhite \underline{0.009} &\doneWhite \underline{-32.4} & 0.267/0.111/0.193 & 0.857 &\doneWhite \textbf{0.008} &\doneWhite \textbf{-37.0} \\
    Substitution & 0.258/0.101/0.201 & 0.846 &\doneWhite 0.010 &\doneWhite -27.2 & 0.249/0.101/0.197 & 0.849 & \doneWhite\underline{0.009} & \doneWhite\underline{-27.6} \\
    DPO   & 0.260/0.109/0.207 & 0.850 & \doneWhite 0.013 & \doneWhite -5.7 & 0.271/0.107/0.213 & 0.863 & \doneWhite 0.012 & \doneWhite -3.6 \\
    Penalty & 0.256/0.110/0.198 & 0.853 &\doneWhite 0.012 &\doneWhite -14.7 & 0.276/0.112/0.207 & 0.863 &\doneWhite 0.009 &\doneWhite -15.7 \\
    Classifier & \textbf{0.274}/0.112/0.207 & 0.859 &\doneWhite 0.011 &\doneWhite -17.7 & \underline{0.279}/0.112/0.209 & 0.862 &\doneWhite 0.011 &\doneWhite -11.0 \\
    \text{IT} & 0.250/0.100/0.192 & 0.844 &\doneWhite 0.012 &\doneWhite -11.0 & \textbf{0.280}/\textbf{0.124}/\textbf{0.216} & 0.860 &\doneWhite 0.010 &\doneWhite -20.5 \\
    \(\text{IT}_{PN_1}\) & 0.263/\underline{0.113}/0.207 & 0.863 &\doneWhite 0.013 &\doneWhite -5.9  & 0.272/0.116/0.212 & 0.867 &\doneWhite 0.012 &\doneWhite -3.2 \\
    \(\text{IT}_{PN_2}\) & \underline{0.265}/\textbf{0.114}/\textbf{0.209} & \textbf{0.866} &\doneWhite 0.012 &\doneWhite -14.0 & 0.273/0.118/\underline{0.215} & \textbf{0.8690} &\doneWhite 0.009 &\doneWhite -26.8 \\
    \(\text{IT}_{NP_1}\) & 0.265/0.112/\underline{0.209} & \underline{0.865} &\doneWhite 0.011 &\doneWhite -17.7 & 0.266/0.115/0.210 & 0.866 &\doneWhite 0.012 &\doneWhite -8.7 \\
    \(\text{IT}_{NP_2}\) & 0.262/0.111/0.205 & 0.862 &\doneWhite \textbf{0.009} &\doneWhite \textbf{-33.8} & 0.275/\underline{0.119}/0.214 & \underline{0.867} &\doneWhite 0.011 &\doneWhite -11.8 \\
    \bottomrule
    \end{tabular}%
    }
    \caption{Results on \textit{\textbf{wikidoc\_patient\_information}}.}
  \label{tab:tab3}
  \vspace{-0.2cm}
\end{table}%
\vspace{-0.2cm}

\begin{figure*}[!ht]\vspace{-0.8cm}
    \centering
    \subfigure[Utility (ROUGE-1) v.s. $S_{Priv}$.]{
        \includegraphics[width=0.475\textwidth]{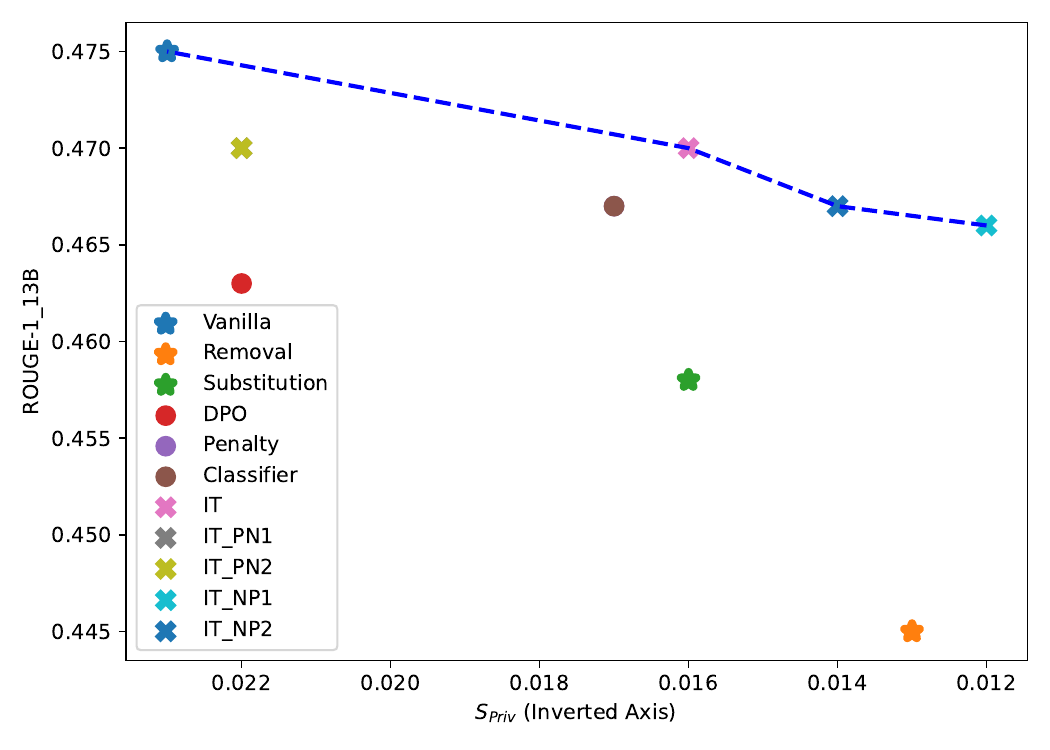}
        \label{fig:pareto_rouge_13b}
    }
    % Adding vertical space between the figures
    \hfill % Adjust this space as needed
    \subfigure[Utility ($S_{BERT}$) v.s. $S_{Priv}$.]{
        \includegraphics[width=0.475\textwidth]{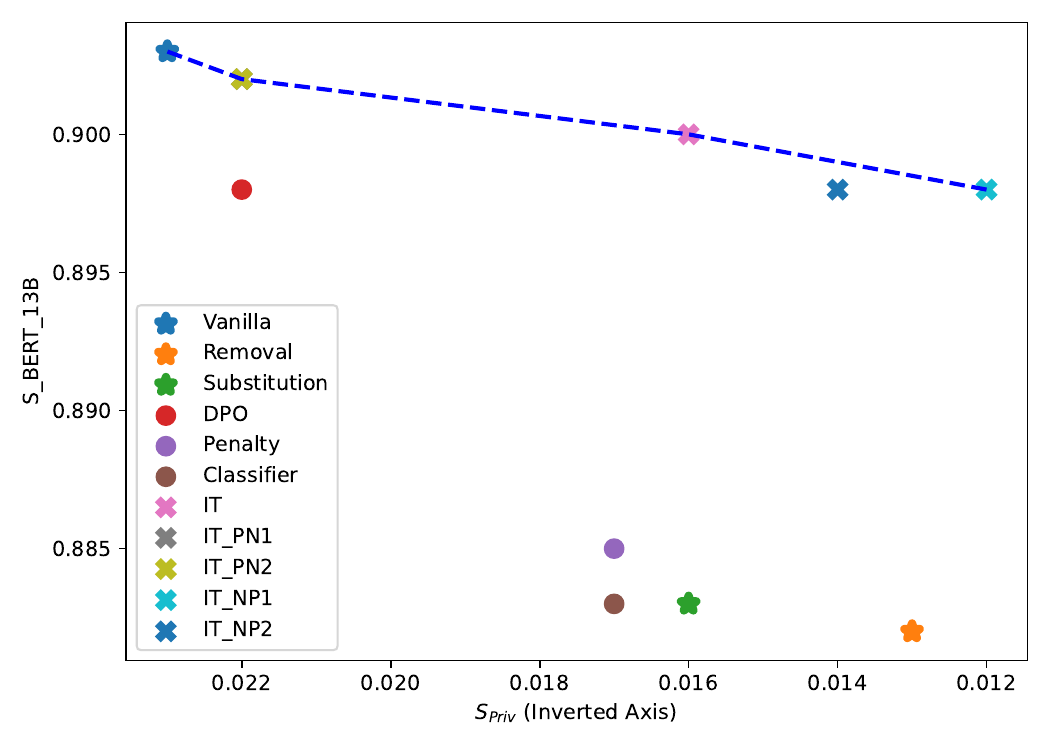}
        \label{fig:pareto_bert_13b}
    }\vspace{-0.3cm}
    \caption{Pareto Frontier.}
    \label{fig:pareto}\vspace{-0.1cm}
\end{figure*}

\begin{table*}[!h]\vspace{-0.35cm}
% \caption{Results on our \textit{\textbf{PQA}} Dataset.}\label{tab:pqa}
\resizebox{\linewidth}{!}{
\begin{tabular}{c|cccc|
>{\columncolor{bubbles}}c 
>{\columncolor{bubbles}}c 
>{\columncolor{bubbles}}c 
>{\columncolor{bubbles}}c 
>{\columncolor{bubbles}}c 
>{\columncolor{bubbles}}c 
>{\columncolor{bubbles}}c 
>{\columncolor{bubbles}}c }\hline
{\color[HTML]{333333} \textbf{Strategy}}                     & {\color[HTML]{333333} \textbf{ROUGE-1}} & {\color[HTML]{333333} \textbf{ROUGE-2}} & {\color[HTML]{333333} \textbf{ROUGE-L}} & {\color[HTML]{333333} \textbf{BERTScore}} & {\color[HTML]{333333} \textbf{$S_{priv:Name}$}} & {\color[HTML]{333333} \textbf{$\Delta_{Name}$}} & {\color[HTML]{333333} \textbf{$S_{priv:Email}$}} & {\color[HTML]{333333} \textbf{$\Delta_{Email}$}} & {\color[HTML]{333333} \textbf{$S_{priv:Address}$}} & {\color[HTML]{333333} \textbf{$\Delta_{Address}$}} & {\color[HTML]{333333} \textbf{$S_{priv:SSN}$}} & {\color[HTML]{333333} \textbf{$\Delta_{SSN}$}} \\\hline
{\color[HTML]{333333} \done Vanilla}                               & {\color[HTML]{333333}\done 0.637}            & {\color[HTML]{333333} \done\underline{0.5743}}  & {\color[HTML]{333333} \done 0.6235}           & {\color[HTML]{2C3A4A} \done \textbf{0.8699}}    & {\color[HTML]{333333}\done 0.0778}                   & {\color[HTML]{333333}\done -}                        & {\color[HTML]{333333}\done 0.0752}                    & {\color[HTML]{333333}\done -}                         & {\color[HTML]{333333}\done 0.0782}                      & {\color[HTML]{333333}\done -}                           & {\color[HTML]{333333}\done 0.0724}                  & {\color[HTML]{333333} \done -}                       \\
{\color[HTML]{333333} Removal}                               & {\color[HTML]{333333} 0.6148}           & {\color[HTML]{333333} 0.5575}           & {\color[HTML]{333333} 0.6115}           & {\color[HTML]{333333} 0.8390}              & {\color[HTML]{333333} 0.0410}                    & {\color[HTML]{333333} -47.30}                    & {\color[HTML]{2C3A4A} \textbf{0.0394}}           & {\color[HTML]{2C3A4A} \textbf{-47.61}}           & {\color[HTML]{333333} 0.0423}                      & {\color[HTML]{333333} -45.91}                      & {\color[HTML]{333333} 0.0419}                  & {\color[HTML]{333333} -42.13}                  \\
{\color[HTML]{333333} Substitution}                          & {\color[HTML]{333333} 0.6291}           & {\color[HTML]{333333} 0.5234}           & {\color[HTML]{333333} 0.6217}           & {\color[HTML]{333333} 0.8576}             & {\color[HTML]{333333} 0.0420}                    & {\color[HTML]{333333} -46.02}                   & {\color[HTML]{333333} 0.0418}                    & {\color[HTML]{333333} -44.41}                    & {\color[HTML]{333333} 0.0446}                      & {\color[HTML]{333333} -42.97}                      & {\color[HTML]{333333} 0.0419}                  & {\color[HTML]{333333} -42.13}                  \\
{\color[HTML]{333333} \(\text{IT}\)}            & {\color[HTML]{333333} \underline{0.6395}}  & {\color[HTML]{333333} 0.5429}           & {\color[HTML]{333333} \underline{0.6253}}  & {\color[HTML]{333333} 0.8686}             & {\color[HTML]{333333} 0.0449}                   & {\color[HTML]{333333} -42.29}                   & {\color[HTML]{333333} 0.0418}                    & {\color[HTML]{333333} -44.41}                    & {\color[HTML]{333333} 0.0449}                      & {\color[HTML]{333333} -42.58}                      & {\color[HTML]{333333} 0.0421}                  & {\color[HTML]{333333} -41.85}                  \\
{\color[HTML]{333333} \(\text{IT}_{PN_1}\)} & {\color[HTML]{2C3A4A} \textbf{0.6497}}  & {\color[HTML]{333333} 0.5591}           & {\color[HTML]{2C3A4A} \textbf{0.6346}}  & {\color[HTML]{333333} \underline{0.8696}}    & {\color[HTML]{2C3A4A} \textbf{0.0395}}          & {\color[HTML]{2C3A4A} \textbf{-49.23}}          & {\color[HTML]{333333} \underline{0.0397}}           & {\color[HTML]{333333} \underline{-47.21}}           & {\color[HTML]{2C3A4A} \textbf{0.0419}}             & {\color[HTML]{2C3A4A} \textbf{-46.42}}             & {\color[HTML]{2C3A4A} \textbf{0.0411}}         & {\color[HTML]{2C3A4A} \textbf{-43.23}}         \\
{\color[HTML]{333333} \(\text{IT}_{PN_2}\)} & {\color[HTML]{333333} 0.6324}           & {\color[HTML]{333333} 0.5569}           & {\color[HTML]{333333} 0.6222}           & {\color[HTML]{333333} 0.869}              & {\color[HTML]{333333} \underline{0.0404}}          & {\color[HTML]{333333} \underline{-48.07}}          & {\color[HTML]{333333} 0.0403}                    & {\color[HTML]{333333} -46.41}                    & {\color[HTML]{333333} \underline{0.0421}}             & {\color[HTML]{333333} \underline{-46.16}}             & {\color[HTML]{333333} \underline{0.0413}}         & {\color[HTML]{333333} \underline{-42.96}}         \\
{\color[HTML]{333333} \(\text{IT}_{NP_1}\)} & {\color[HTML]{333333} 0.6321}           & {\color[HTML]{333333} 0.5740}            & {\color[HTML]{333333} 0.6234}           & {\color[HTML]{333333} 0.8605}             & {\color[HTML]{333333} 0.0411}                   & {\color[HTML]{333333} -47.17}                   & {\color[HTML]{333333} 0.0412}                    & {\color[HTML]{333333} -45.21}                    & {\color[HTML]{333333} 0.0431}                      & {\color[HTML]{333333} -44.88}                      & {\color[HTML]{333333} 0.0414}                  & {\color[HTML]{333333} -42.82}                  \\
{\color[HTML]{333333} \(\text{IT}_{NP_2}\)} & {\color[HTML]{333333} 0.6335}           & {\color[HTML]{2C3A4A} \textbf{0.5761}}  & {\color[HTML]{333333} 0.6201}           & {\color[HTML]{333333} 0.8657}             & {\color[HTML]{333333} 0.0406}                   & {\color[HTML]{333333} -47.81}                   & {\color[HTML]{333333} 0.0408}                    & {\color[HTML]{333333} -45.74}                    & {\color[HTML]{333333} 0.0412}                      & {\color[HTML]{333333} -47.31}                      & {\color[HTML]{333333} 0.0416}                  & {\color[HTML]{333333} -42.54}         \\\hline        
\end{tabular}
}
\caption{Results on our \textit{\textbf{PQA}} Dataset.}\label{tab:pqa}\vspace{-0.5cm}
\end{table*}

Selective forgetting constraints in models may inadvertently alter existing knowledge, leading to token alterations for PIIs and possibly distorting original information, slightly reducing performance in some datasets. The ``Classifier'' approach offers moderate privacy protection results, reflecting the challenge in training contextual classifiers. DPO starts with Vanilla tuning (SFT) without privacy measures, then fine-tunes for PII concealment without a reward model. While DPO boosts privacy through preference-based tuning, its effectiveness is limited, often needing a larger dataset of user preferences and facing reward hacking issues.

Experiments show that instruction tuning with examples, using instructions and examples for fine-tuning, achieves a good balance between performance, privacy, information preservation, and alignment with human preferences. This method, letting the model ``see'' and ``learn'' from both preferred and undesired examples, helps in aligning the model. It enables the model to understand what information to withhold, highlighting the potential of LLMs in privacy protection learning.

We also plot the Pareto frontier in Figure \ref{fig:pareto_rouge_13b} and \ref{fig:pareto_bert_13b} to evaluate both utility and privacy preservation on \textit{medical\_flashcards} dataset for \texttt{LLaMA2-7B} and \texttt{LLaMA2-13B}, respectively. More results are reported in Appendix \ref{app:pareto}. It is evident that the instruction-based approaches consistently align with the Pareto frontier (\emph{IT} methods constitute the border of the frontier). Such a phenomenon indicates that employing instructions supplemented by both positive and negative examples achieves the optimal trade-off between performance (utility) and privacy protection of PIIs.

% \begin{figure}[htbp]
% \centering  
% \begin{subfigure}[b]{0.45\textwidth}
%         \centering
% \includegraphics[width=0.4\linewidth]{figures/pareto/medical_flashcards_13_ROUGE-1_13B.pdf}
%   % \vspace{-0.3cm}
%   \caption{Pareto Frontier: Utility (ROUGE-1) v.s. $S_{Priv}$.}
%   \label{fig:pareto_rouge_13b}
% \end{subfigure}
%     \hfill
% % \vspace{-0.3cm}
% % \begin{figure}[htbp]
% % \centering  
% \begin{subfigure}[b]{0.45\textwidth}
%         \centering
%         \includegraphics[width=0.4\linewidth]{figures/pareto/medical_flashcards_13_S_BERT_13B.pdf}
%   % \vspace{-0.3cm}
%   \caption{Pareto Frontier: Utility ($S_{BERT}$) v.s. $S_{Priv}$.}
%   \label{fig:pareto_bert_13b}
%   \end{subfigure}
%   \caption{Main figure with two subfigures}
% \end{figure}

\vspace{-0.15cm}
\subsection{Performance on Different Types of PIIs}\label{app:att}
\vspace{-0.15cm}
To validate the performance of our approaches on different types of PIIs, we have conducted further experiments on the newly synthesized dataset. The dataset, named Privacy QA (PQA) Dataset, was synthesized using GPT-4. The PQA dataset contains a wider range of entities, including Names, Emails, Addresses, and SSNs. PQA is accessible at the anonymous link\footnote{\url{https://github.com/Yijia-Xiao/PrivacyMind/ft_datasets/data/PQA.csv}}. The categorization helps assess the protection effectiveness for each PII type. For instance, SSN leaks are generally more critical than name leaks. We performed experiments on the Privacy QA dataset, evaluating the protection ratios across these PII categories respectively. The evaluation is performed on LLaMA2-7B and results are provided in Table \ref{tab:pqa}. The results show that the instruction tuning approaches can well protect different types of PIIs while providing good knowledge injections.

\vspace{-0.2cm}
\section{Conclusion}
\vspace{-0.15cm}
In this paper, we present a comprehensive exploration of strategies for fine-tuning Large Language Models (LLMs) to incorporate domain-specific knowledge while upholding data privacy, particularly in safeguarding sensitive Personally Identifiable Information (PII). We introduced the novel concept of Contextual Privacy Protection Language Models (CPPLMs) and provided a theoretical analysis to guide model design. Our extensive experiments underscore the effectiveness of our approach, with instruction-based tuning emerging as a promising method to simultaneously protect private data and enhance the model's knowledge. 
This study highlights the potential for LLMs to serve as adept privacy protection learners, bridging the gap between domain-specific expertise and data privacy. As LLMs continue to play a pivotal role in natural language understanding and generation, our findings contribute to advancing their utility in privacy-sensitive applications.

\clearpage
\newpage

% \section*{Limitations}

\section*{Limitations}

CPPLM explores privacy preservation in large language models. It is important to note that in our dataset, personally identifiable information (PII) is identified using the scrubadub toolkit. Such a tagging process may not fully represent real-world deployment scenarios, where users can customize privacy preferences. Companies and data owners can employ the CPPLM pipeline to teach language models contextual privacy from annotated positive-negative pairs. Since there is no universal rule for detecting PIIs, privacy definitions vary across scenarios. Therefore, our focus is on demonstrating the language model's ability to learn contextual PII. For instance, a clinical company wanting to protect specific PIIs can annotate datasets and follow our proposed method. Even end-users may define what PII means in their data's context during language model tuning or training. In summary, the CPPLM pipeline is versatile and adaptable to various privacy-related scenarios and tasks, such as detoxifying language models.

\section*{Acknowledgments}
This work was supported by NSF grants \#2200274, \#2106859, \#2312501, NIH grants \#U54HG012517, \#U24DK097771.

\bibliography{main}
\bibliographystyle{acl_natbib}

\newpage
\clearpage
\appendix
\appendix

 \onecolumn
 \begin{center}
    {\Large \textbf{Appendix: Large Language Models Can Be Contextual Privacy Protection Learners}}
 \end{center}
 %\twocolumn
 
\section{Notations}
Important notations used in the paper are included in Table. \ref{tab:notation}.
\begin{table*}[!th]
\small
\centering
\caption{Notations used in this paper.}
\label{tab:notation}
\begin{tabular}{c|l}
\toprule
\textbf{Notation} & \textbf{Description} \\
\midrule
\( w_i, \textbf{w}_i \) & a token and its contextualized embedding \\
\( s \) & a natural language sequence \\
\( D = \{s\} \) & Fine-tuning dataset \\ \midrule
\( T \) & Annotation \\
\( n \) & Maximum sequence length \\
\( \Theta_{n} \) & Set of n-grams associated with PII \\ \midrule
\( R \) & Removed sequence: \( R = (r_0, r_1, ..., r_{k-1}) \) \\
\( r_{i-1} \) & i-th token with \( p_i = 0 \) in sequence \( R \) \\
\( C \) & Cleaned sequence: \( C = (c_0, c_1, ..., c_{n-1}) \) \\
\( y_i \) & Token \( c_i \) if \( p_i = 0 \), or the special token \( u \) if \( p_i = 1 \) \\
\( u \) & Special token added to the vocabulary (e.g., \emph{unk} for LLaMA2) \\
$\mathbb{P}(\cdot)$ & Probability \\
\bottomrule
\end{tabular}
\vspace{-0.2cm}
\end{table*}

\subsection{Detailed Datasets Description} \label{app:dataset}
Table. \ref{tab:stat} shows more details about datasets: \(S\) denotes the size of the train/test set and \(L_Q\)/\(L_A\) denotes the average length (number of tokens) of the question/answer fields.
(1) \verb|pii-medical_flashcards| with 28861 training and 5093 testing samples; (2) \verb|pii-wikidoc| with 8500 training and 1500 testing samples; (3) \verb|pii-wikidoc_patient_information| with 5050 training and 891 testing samples.

\begin{table*}[htbp]
  \centering
  % \captionsetup[table]{skip=4pt}
  \caption{Statistics of datasets}
    \begin{tabular}{lcccccc}
    \toprule
    \multirow{2}[4]{*}{\textbf{Dataset}} & \multicolumn{3}{c}{\textbf{Train}} & \multicolumn{3}{c}{\textbf{Test}} \\
\cmidrule{2-7}          & \(|S|\)  & \(L_Q\) & \(L_A\) & \(|S|\)  & \(L_Q\) & \(L_A\) \\
    \midrule
    medical-flashcards & 28861 & 14.59 & 14.36 & 5093  & 53.64 & 52.74 \\
    medical-wikidoc & 8500  & 9.88  & 9.67  & 1500  & 132.04 & 136.60 \\
    wikidoc-patient-information & 5050  & 8.15  & 8.04  & 891   & 73.40 & 71.10 \\
    \bottomrule
    \end{tabular}%
  \label{tab:stat}%
\end{table*}%

\subsection{Metrics} \label{app:metrics}

\paragraph{ROUGE (Recall-Oriented Understudy for Gisting Evaluation)}
We adopt the popularly used ROUGE-1, ROUGE-2, ROUGE-L \citep{lin-2004-rouge} and BERTScore \citep{bert-score} to evaluate the answer quality in the testing phase. Here we give a detailed definition of these scores. We denote the set of tokens from the generated text as \(G\), and the set of tokens from the reference text as \(R\). The number of overlapping unigrams between \(G\) and \(R\) as \(O_1(G, R)\), and the number of overlapping bigrams between \(G\) and \(R\) as \(O_2(G, R)\).
The total number of unigrams in \(R\) as \(U(R)\) and the total number of bigrams in \(R\) as \(B(R)\). The longest common subsequence (LCS) between \(G\) and \(R\) as \(L(G, R)\).

\noindent \textbf{ROUGE-1:}
\[
\text{ROUGE-1} = \frac{O_1(G, R)}{U(R)}
\]
\textbf{ROUGE-2:}
\[
\text{ROUGE-2} = \frac{O_2(G, R)}{B(R)}
\]
\textbf{ROUGE-L:}
\[
\text{ROUGE-L} = \frac{L(G, R)}{\max(|G|, |R|)}
\]

\paragraph{BERTScore}
\small{
\begin{align*}
E & : \text{BERT encoder or model} \\
E(G) & : \text{Embedding of the entire sequence}\\ 
\quad&\text{ of the generated text } G, \text{ produced by } E \\
E(R) & : \text{Embedding of the entire sequence }\\ 
\quad&\text{of the reference text } R, \text{ produced by } E \\
c(E(G), E(R)) & : \text{Cosine similarity between the }\\ 
\quad&\text{sequence embeddings } E(G) \text{ and } E(R)
\end{align*}
}\normalsize

Then, the BERTScore between a generated text \( G \) and a reference text \( R \) at the sequence level is defined as:
\[
\text{BERTScore}(G, R) = c(E(G), E(R))
\]

Here, the BERT model \( E \) encodes the entire sequences \( G \) and \( R \) into their respective embeddings, and then we compute the cosine similarity between these sequence embeddings to obtain the BERTScore.

\section{Additional Related Work}

\stitle{Pretraining with Preferences.}
Another solution is to maintain the content, but use redesigned loss/conditional tags to control the information injected into the LLMs. 
Pretraining with conditional human preference scores can offer a \revise{Pareto-optimal} and simple approach to reduce the undesirable content by up to an order of magnitude. \citet{Korbak2023PretrainingLM} compared with the classical pretraining approach. While pretraining LLMs conditioned under annotation scores can offer better performance in the human preferences aspect. Since human preferences are injected into the models during the pretraining stage, the models are biased toward those preferences once they are trained. With the expanding size of LLMs, they become increasingly resistant to forgetting their training data \citep{Carlini2022QuantifyingMA, Vu2022OvercomingCF, Ramasesh2022EffectOS, Korbak2023PretrainingLM}. In other words, pretraining large language models conditioned under preference score sacrifices some flexibility. Still, it is undeniable that it can provide much better alignment with human preferences compared with the classical pretraining schema.

\section{Illustration of Vanilla Tuning and Corpus Curation}
\label{sec:figvanilla}
\begin{figure*}[htbp]
    \centering
    \subfigure[Classical Tuning]{
        \includegraphics[width=0.7\linewidth]{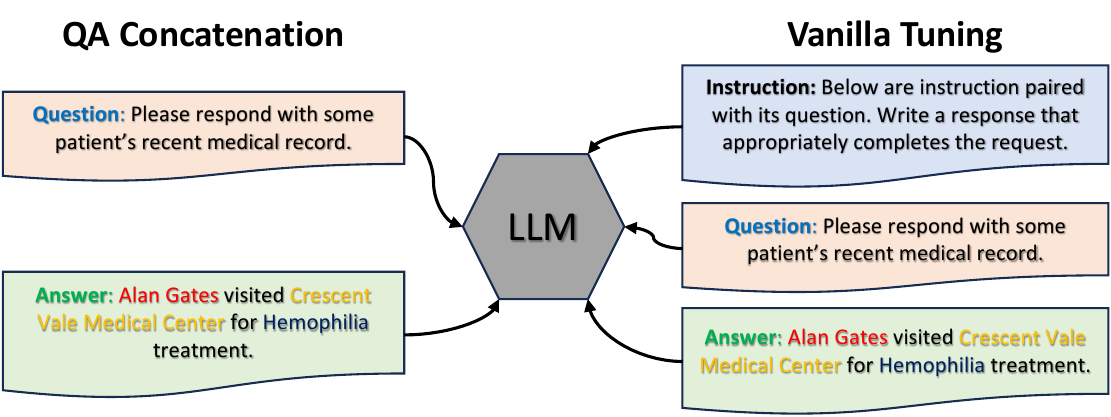}
        \label{fig:vanilla}
    }
    %\hfill
    \subfigure[Corpus Curation]{
        \includegraphics[width=0.7\linewidth]{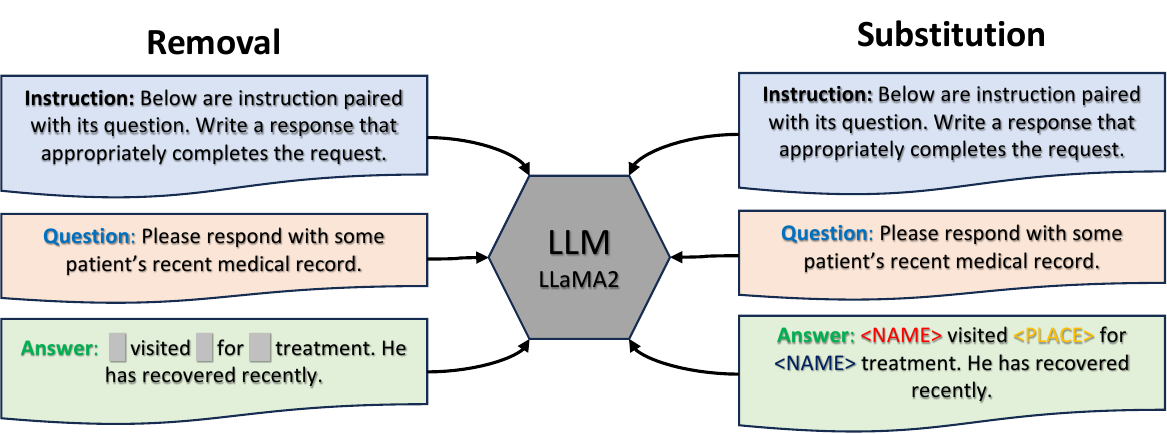}
        \label{fig:corpus}
    }

    \caption{Vanilla, Removal, Substitution.}
    \label{fig:tuning}
\end{figure*}

This section gives an illustration of Vanilla Tuning (Figure. \ref{fig:vanilla}) and Corpus Curation (Figure. \ref{fig:corpus}).

\section{Experiment Details.}
\label{sec:appExp}

\subsection{Hardware and Implementations}
In this paper, we implemented our method on two Linux servers with 4 NVIDIA A100 GPUs, each with 80GB of memory. The CUDA version is 12.2 and the Driver version is 535.54.03. We used Python 3.10.12 and Pytorch 2.0.1~\citep{paszke2019pytorch} to construct our project. The fine-tuning of LLaMA models takes ~20 hours on average.

\subsection{Dataset and Hyperparameters}
In our experiments, we use grid search to obtain the best performance. We provide all of the hyperparameters as well as their configurations in the following:

\begin{itemize}
    \item \textbf{Dataset.} For training, we sub-sampled 85\% from the three datasets. The performance of each method is evaluated on the remaining 15\% of data. Dataset details can be found in Table. \ref{tab:stat}.

    \item \textbf{Hyperparameters.} For the parameter optimizer, we chose \verb|AdamW| with \verb|weight_decay| set to 0. The learning rate is set to \(1e^{-4}\). We use the \verb|StepLR| learning rate scheduler with \verb|gamma| set to 0.85. Epochs and Batch Size: The number of fine-tuning epochs is set to 5, and the batch size is set to 64.
\end{itemize}

\subsection{Comparison with Other Methods}\label{app:comparison}

In this section, we compare our proposed methods against several privacy-preserving techniques, including Vanilla, Removal, Substitution, and Private Transformer, as well as our IT\_PN and IT\_NP strategies. The comparison is made on the PQA dataset in terms of both performance (ROUGE and BERTScore metrics) and privacy protection across multiple Personally Identifiable Information (PII) categories such as Name, Email, Address, and SSN.

As shown in Table \ref{tab:comparison}, our methods demonstrate competitive performance in terms of utility, while also providing notable improvements in privacy protection. The IT\_PN method, in particular, consistently outperforms the baseline strategies in terms of reducing privacy leakage, achieving a reduction of up to 49.26\% in the \textit{S\_priv:Name} metric compared to the Vanilla model. Furthermore, IT\_PN maintains a high level of utility, reflected in its ROUGE and BERTScore values, which are comparable to or better than those of the Private Transformer approach.

\begin{table*}[htbp]
\centering

\resizebox{\linewidth}{!}{
\begin{tabular}{c|cccc|
>{\columncolor{bubbles}}c 
>{\columncolor{bubbles}}c 
>{\columncolor{bubbles}}c 
>{\columncolor{bubbles}}c 
>{\columncolor{bubbles}}c 
>{\columncolor{bubbles}}c 
>{\columncolor{bubbles}}c 
>{\columncolor{bubbles}}c }\hline
\textbf{Strategy}                     & \textbf{ROUGE-1} & \textbf{ROUGE-2} & \textbf{ROUGE-L} & \textbf{BERTScore} & \textbf{$S_{priv:Name}$} & \textbf{$\Delta_{Name}$} & \textbf{$S_{priv:Email}$} & \textbf{$\Delta_{Email}$} & \textbf{$S_{priv:Address}$} & \textbf{$\Delta_{Address}$} & \textbf{$S_{priv:SSN}$} & \textbf{$\Delta_{SSN}$} \\\hline
\textbf{Vanilla}                               & 0.3342           & 0.2174           & 0.3297           & 0.8162             & 0.1082                    & -                         & 0.1024                    & -                         & 0.1103                      & -                           & 0.0992                  & -                       \\
\textbf{Removal}                               & 0.2947           & 0.2071           & 0.3092           & 0.7983             & 0.0558                    & -48.43\%                  & 0.0536                    & -47.66\%                  & 0.0573                      & -48.05\%                    & 0.0568                  & -42.75\%                  \\
\textbf{Substitution}                          & 0.2983           & 0.2073           & 0.3173           & 0.8012             & 0.0572                    & -47.13\%                  & 0.0569                    & -44.43\%                  & 0.0586                      & -46.87\%                    & 0.0568                  & -42.75\%                  \\
\textbf{Private Transformer}                   & 0.3172           & 0.2096           & 0.3192           & \underline{0.8119} & \underline{0.0551}         & \underline{-49.08\%}      & 0.0554                    & -45.90\%                  & \underline{0.0572}          & \underline{-48.14\%}        & \underline{0.0569}      & \underline{-42.65\%}      \\
\textbf{IT\_PN}                                & \textbf{0.3273}  & \underline{0.2112} & \underline{0.3221} & 0.8101             & \textbf{0.0549}            & \textbf{-49.26\%}         & \underline{0.0551}         & \underline{-46.19\%}       & \textbf{0.0570}             & \textbf{-48.32\%}           & 0.0570                  & -42.55\%                  \\
\textbf{IT\_NP}                                & \underline{0.3261} & \textbf{0.2162}  & \textbf{0.3252}  & \textbf{0.8132}    & 0.0563                    & -47.97\%                  & 0.0561                    & -45.21\%                  & 0.0575                      & -47.87\%                    & 0.0573                  & -42.24\%                  \\\hline
\end{tabular}
}
\caption{Performance and Privacy Metrics Comparison on the \textit{\textbf{PQA}} Dataset}
\label{tab:comparison}
\end{table*}

\subsection{Pareto Frontier of Utility and Privacy Protection}\label{app:pareto}
\begin{figure*}[htbp]
\centering  \includegraphics[width=\linewidth]{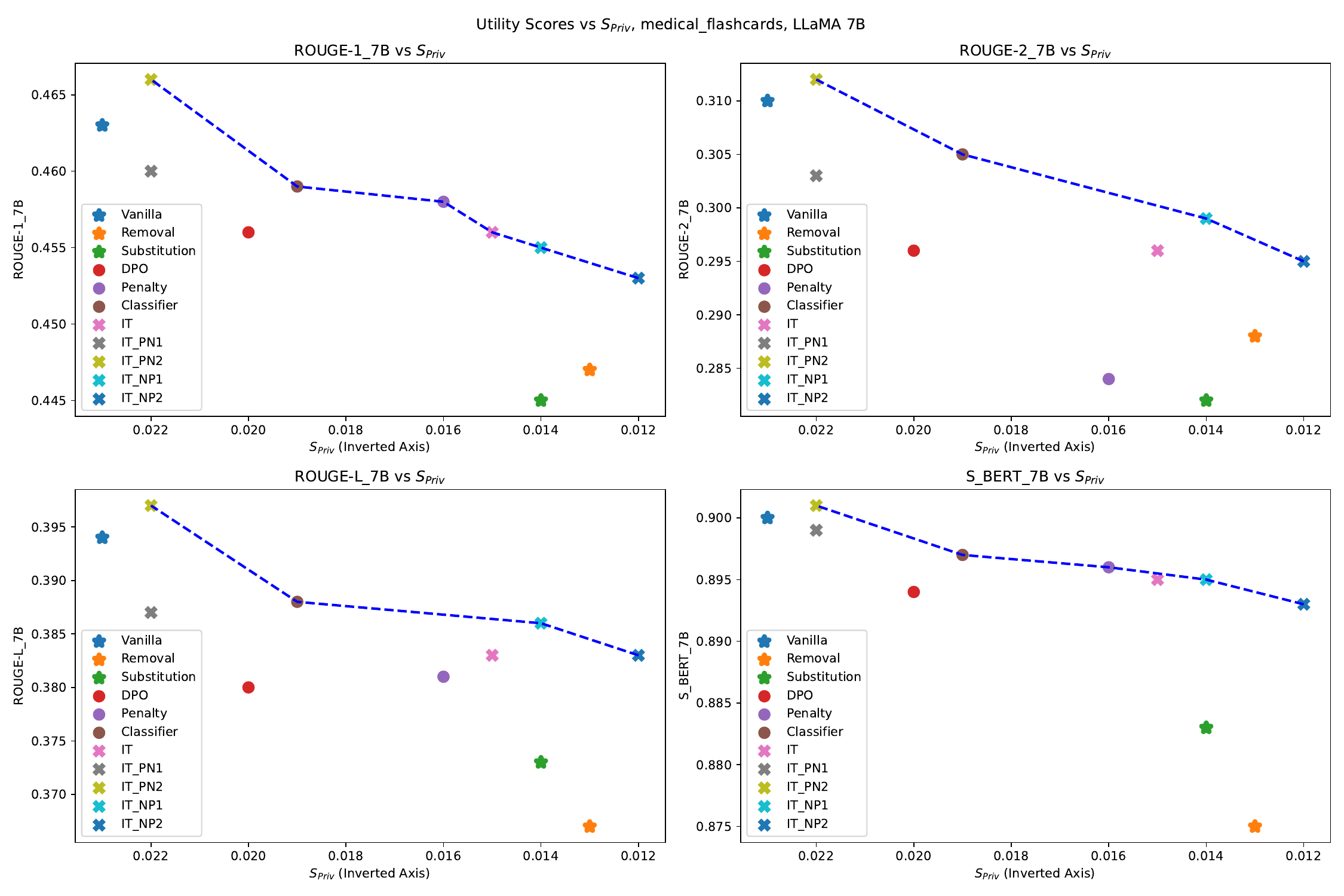}
  \caption{Pareto Frontier on medical\_flashcards, LLaMA2-7B}
  \label{fig:medical_7b}
\end{figure*}

\begin{figure*}[htbp]
\centering  \includegraphics[width=\linewidth]{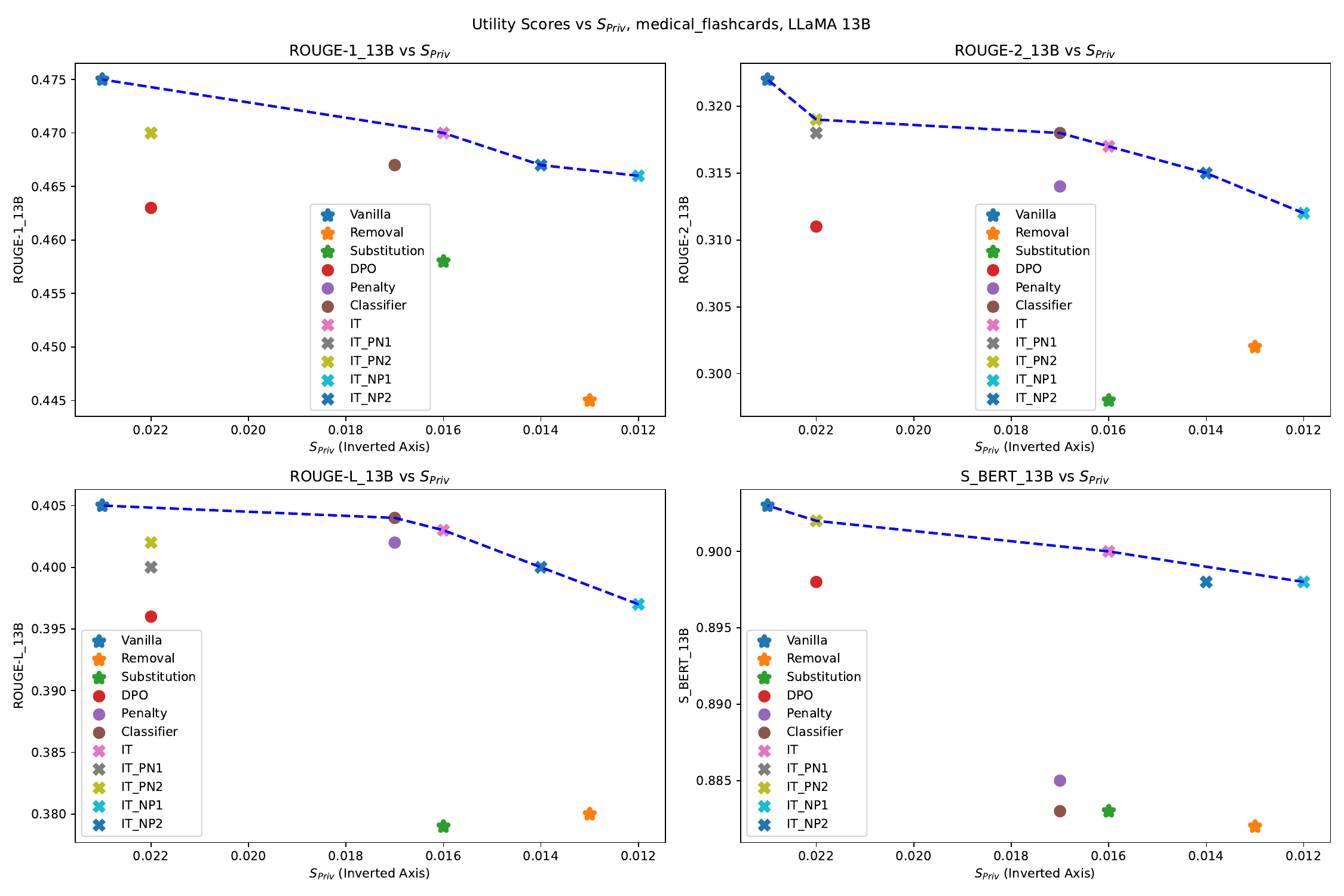}
  % \caption{Pareto Frontier: Utility v.s. $S_{Priv}$}
  \caption{Pareto Frontier on medical\_flashcards, \texttt{LLaMA2-13B}}
  \label{fig:medical_13b}
\end{figure*}

\begin{figure*}[htbp]
\centering  \includegraphics[width=\linewidth]{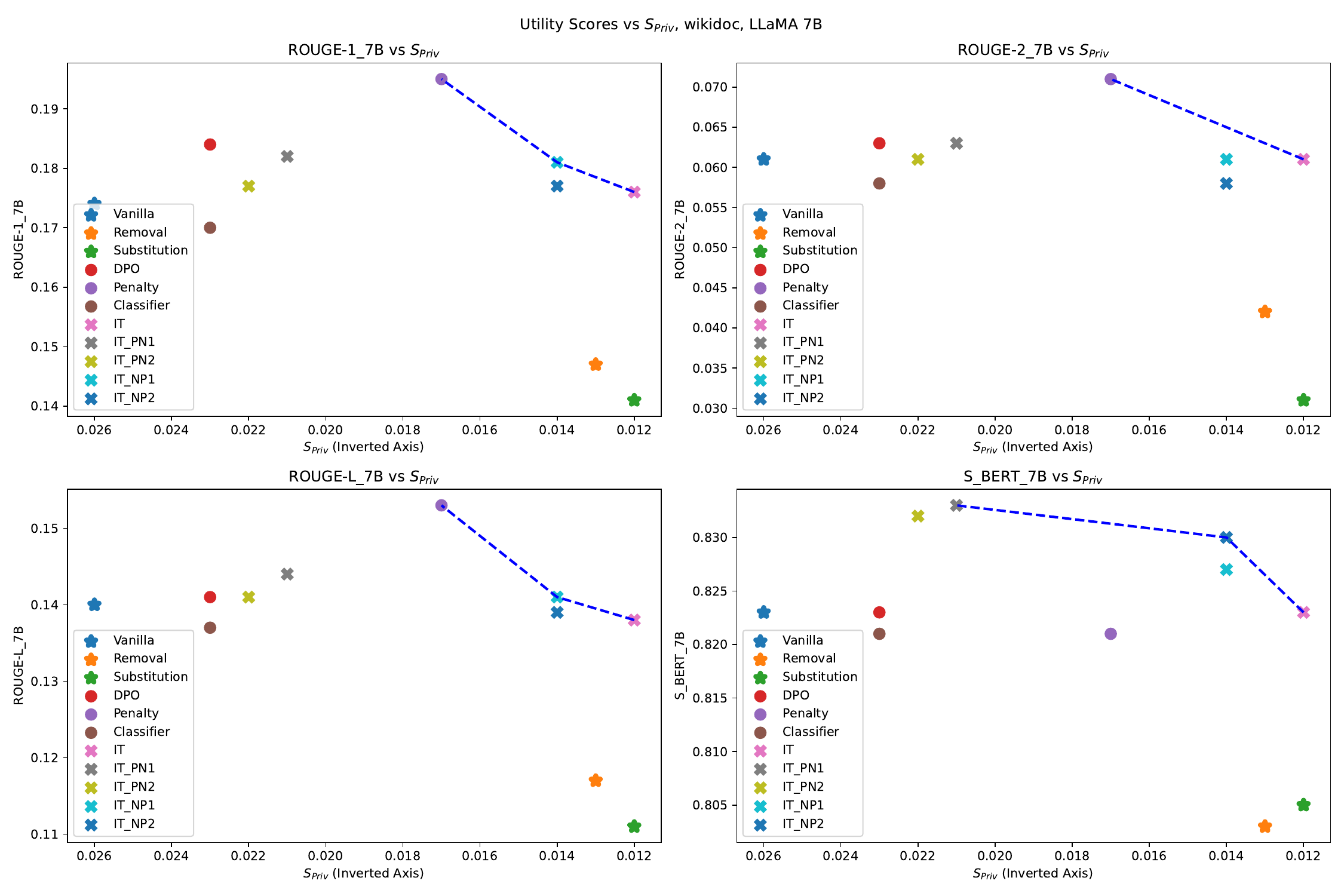}
  \caption{Pareto Frontier on wikidoc, \texttt{LLaMA2-7B}}
  \label{fig:wikidoc_7b}
\end{figure*}

\begin{figure*}[htbp]
\centering  \includegraphics[width=\linewidth]{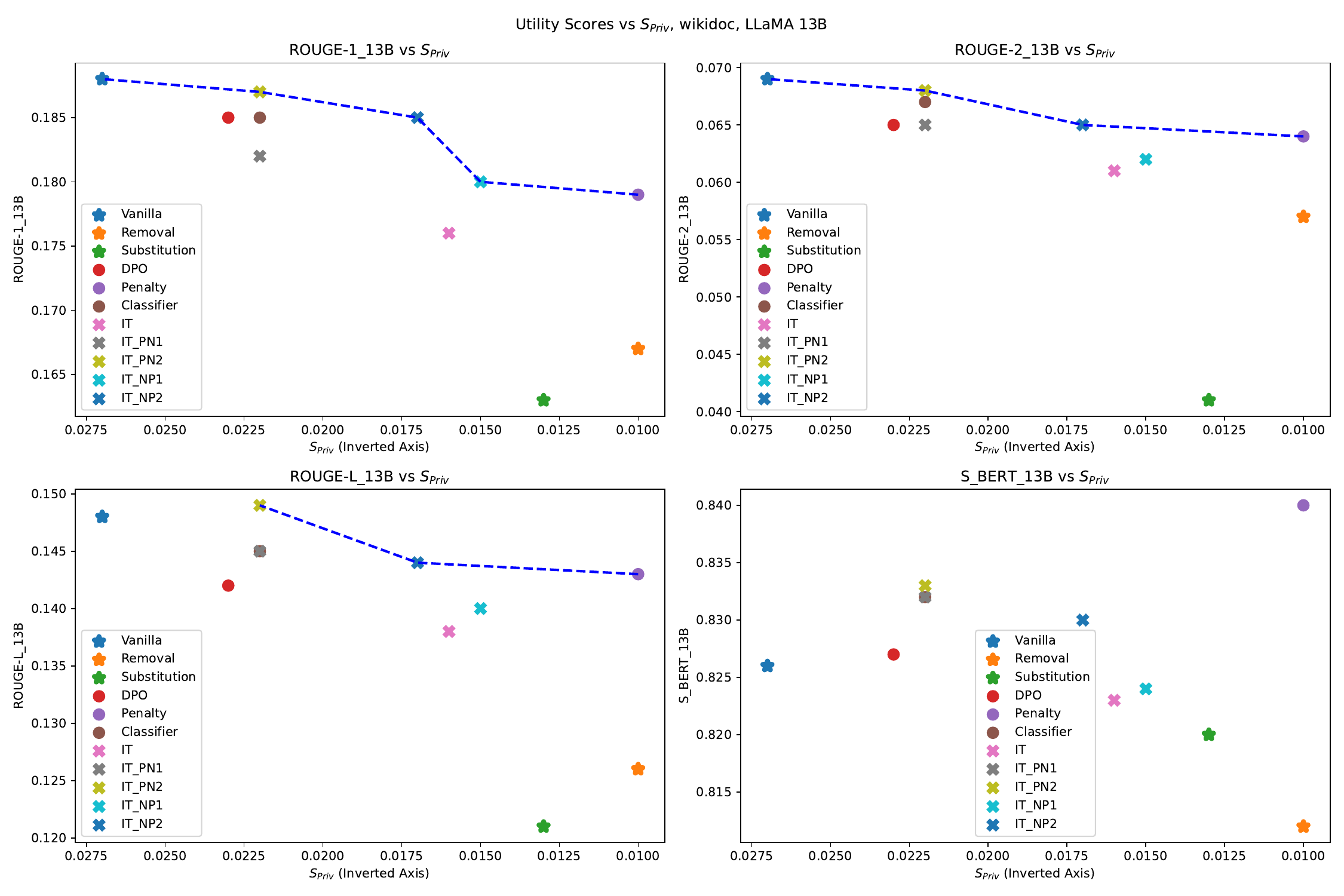}
  % \caption{Pareto Frontier: Utility v.s. $S_{Priv}$}
  \caption{Pareto Frontier on wikidoc, \texttt{LLaMA2-13B}}
  \label{fig:wikidoc_13b}
\end{figure*}

\begin{figure*}[htbp]
\centering  \includegraphics[width=\linewidth]{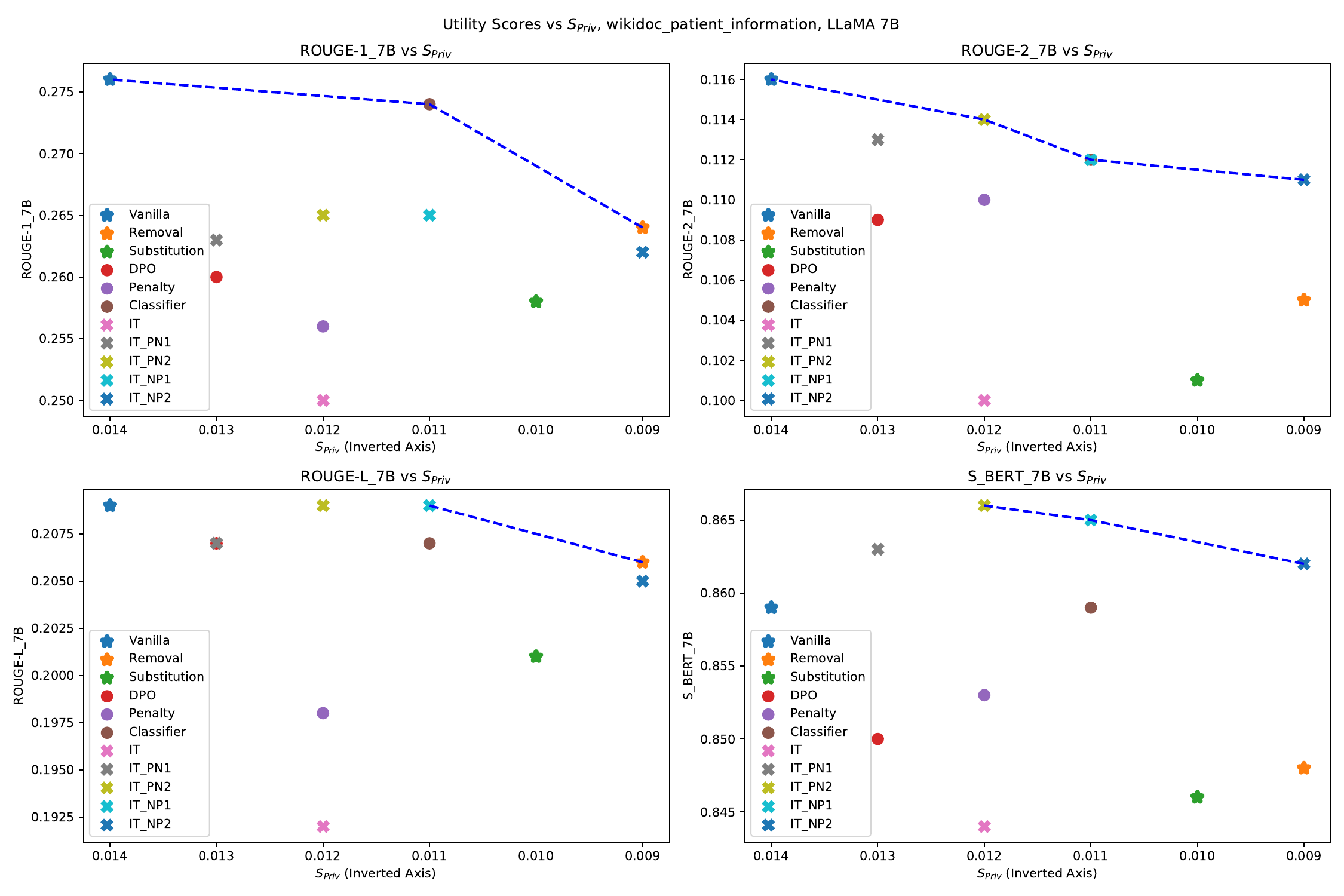}
  \caption{Pareto Frontier on wikidoc\_patient\_information, \texttt{LLaMA2-7B}}
  \label{fig:patient_7b}
\end{figure*}

We also report the pareto frontier of Utility and Privacy Protection in Figure \ref{fig:medical_7b}, \ref{fig:medical_13b}, \ref{fig:wikidoc_7b}, \ref{fig:wikidoc_13b}, \ref{fig:patient_7b}  and \ref{fig:patient_7b}, respectively, to evaluate both performance and privacy preservation. It is obvious that the instruction-based approaches consistently align with the Pareto frontier (\emph{IT} methods constitute the border of the frontier). Such a phenomenon indicates that employing instructions supplemented by both positive and negative examples achieves the optimal trade-off between performance (utility) and privacy protection of PIIs. The outcomes strongly support our position that LLMs can be good contextual privacy protection learners.

\begin{figure*}[htbp]
\centering  \includegraphics[width=\linewidth]{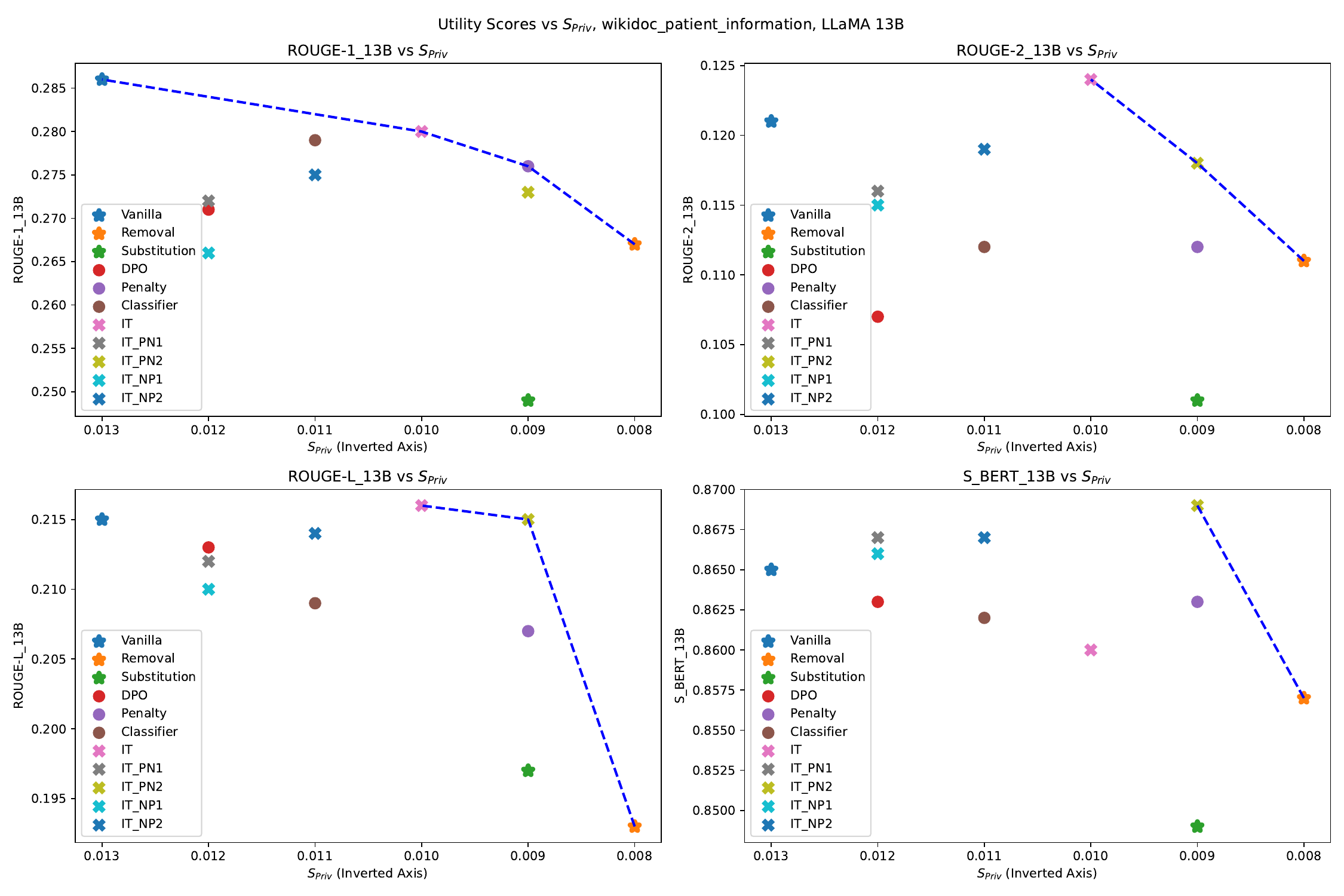}
  % \caption{Pareto Frontier: Utility v.s. $S_{Priv}$}
  \caption{Pareto Frontier on wikidoc\_patient\_information, \texttt{LLaMA2-13B}}
  \label{fig:patient_13b}
\end{figure*}

\subsection{Curve of Knowledge Injection and PII Leakage vs. Learning Process}
\label{sec:learning_process}
\begin{figure}[htbp]
% \vspace{-0.35cm}
    \centering
    \subfigure[ROUGE/BERTScore Curve]{\includegraphics[width=0.47\linewidth]{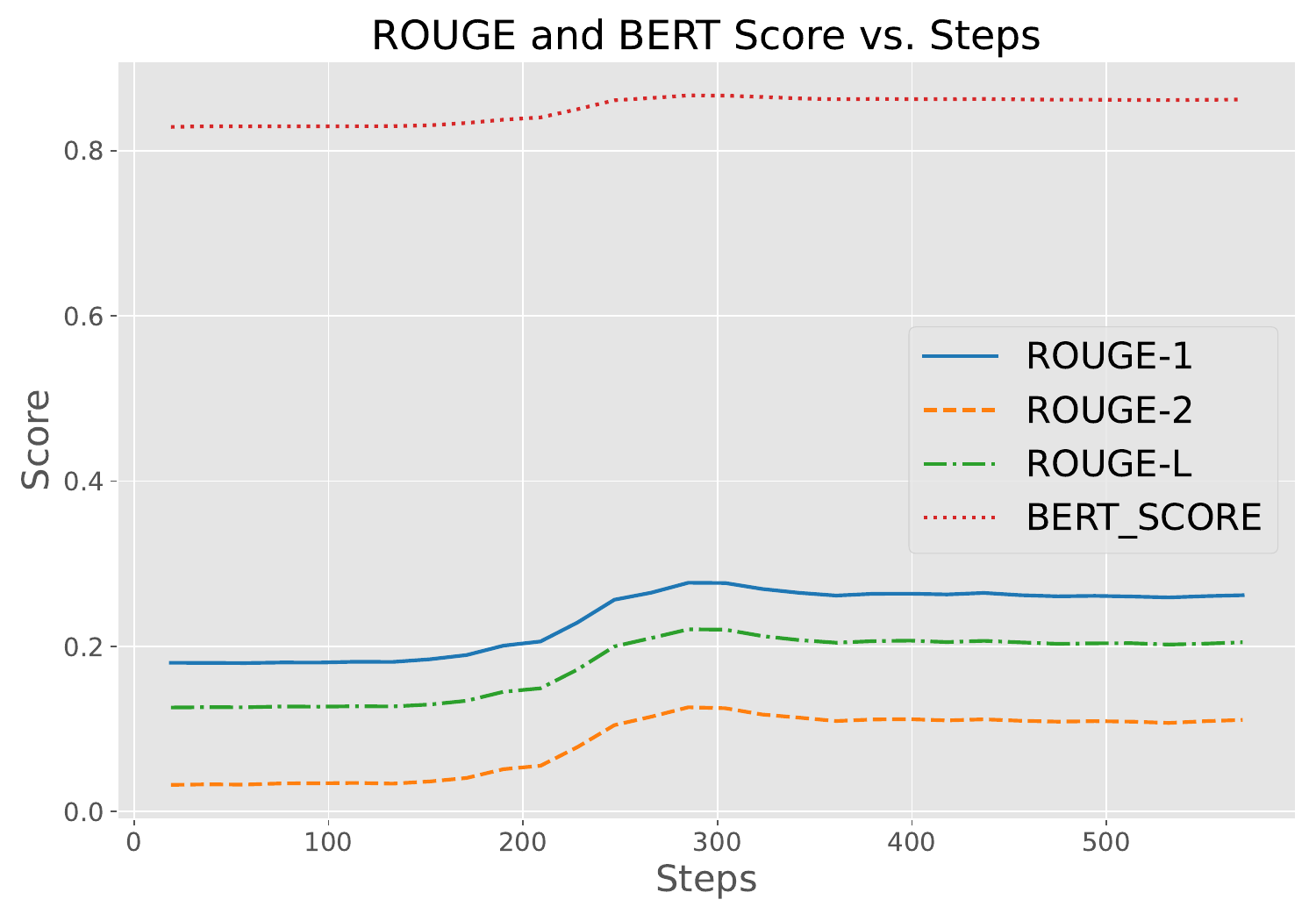}
        \label{fig:metrics}
    }
    %\hfill
    \subfigure[Privacy Score Curve]{
        \includegraphics[width=0.48\linewidth]{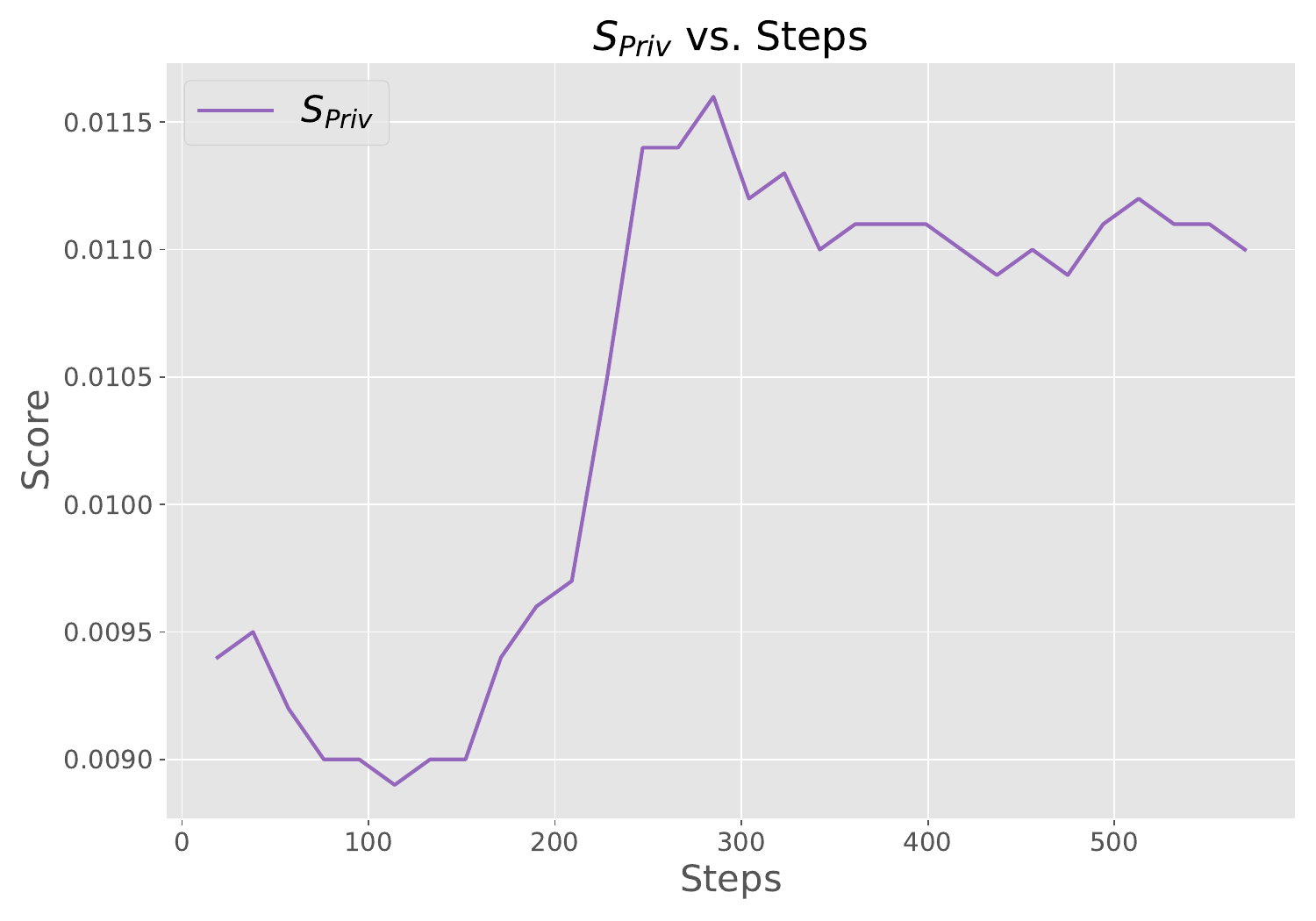}
        \label{fig:priv}
    }
    \caption{ROUGE, BERTScore, and $S_{\text{Priv}}$ vs. Steps}
    \label{fig:curve}
    % \vspace{-0.5cm}
\end{figure}

In this section, we analyze the ROUGE, BERTScore, and Privacy Leakage Score concerning the training steps. We aim to assess whether our two primary learning objectives are effectively achieved throughout the training process. 
Initially, in Figure. \ref{fig:curve} that visualizes the training of $IT_{PN_1}$, we observe that as the LM undergoes the training process, we witness a notable trend: the injection of knowledge into the LM steadily increases. This infusion of knowledge corresponds to a progressive rise in both ROUGE and BERTScore, ultimately leading to a stabilization, or convergence, of these metrics. Simultaneously, the Privacy Leakage Score exhibits an intriguing behavior. At the outset of the learning process, it experiences an upward trajectory. This ascent is a direct consequence of the LM ingesting more knowledge, including private tokens, inadvertently learning about sensitive information. However, as training continues, a pivotal shift occurs. The LM's instruction to conceal privacy-related information gradually takes effect, resulting in a discernible decrease in the Privacy Leakage Score.
In summary, Figure. \ref{fig:curve} offers a compelling visualization of the evolving relationship between knowledge injection, linguistic performance (ROUGE/BERTScore), and privacy protection ($S_{\text{Priv}}$) as the LM matures throughout its training steps. It underscores the dynamic equilibrium between knowledge acquisition and safeguarding sensitive data, emphasizing the importance of a well-orchestrated learning process to achieve both objectives.

\begin{figure}[htbp]
% \vspace{-0.3cm}
    \centering
    \subfigure[ROUGE/BERTScore Curve]{\includegraphics[width=0.47\linewidth]{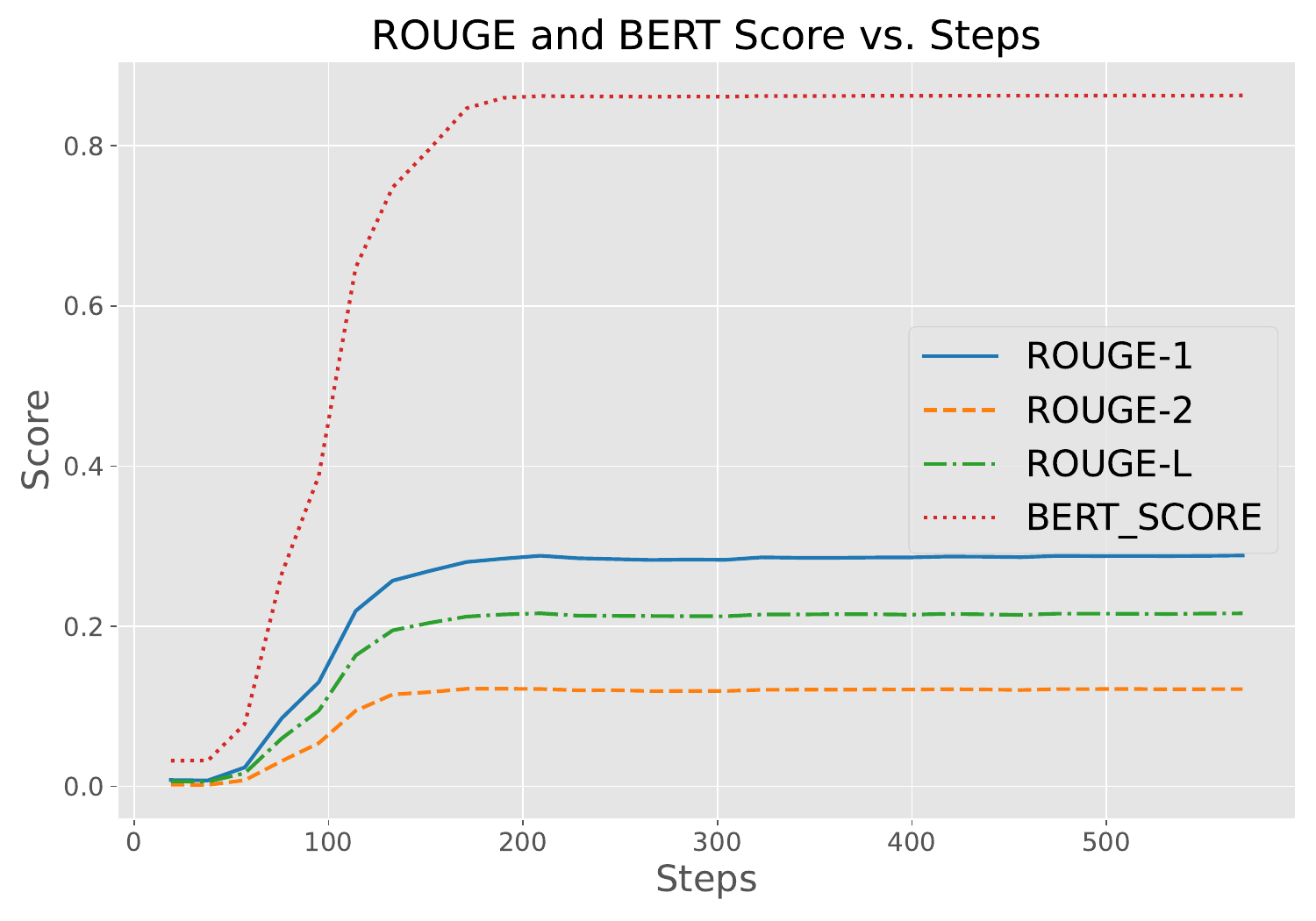}
        \label{fig:orig_metrics}
    }
    %\hfill
    \subfigure[Privacy Score Curve]{
        \includegraphics[width=0.48\linewidth]{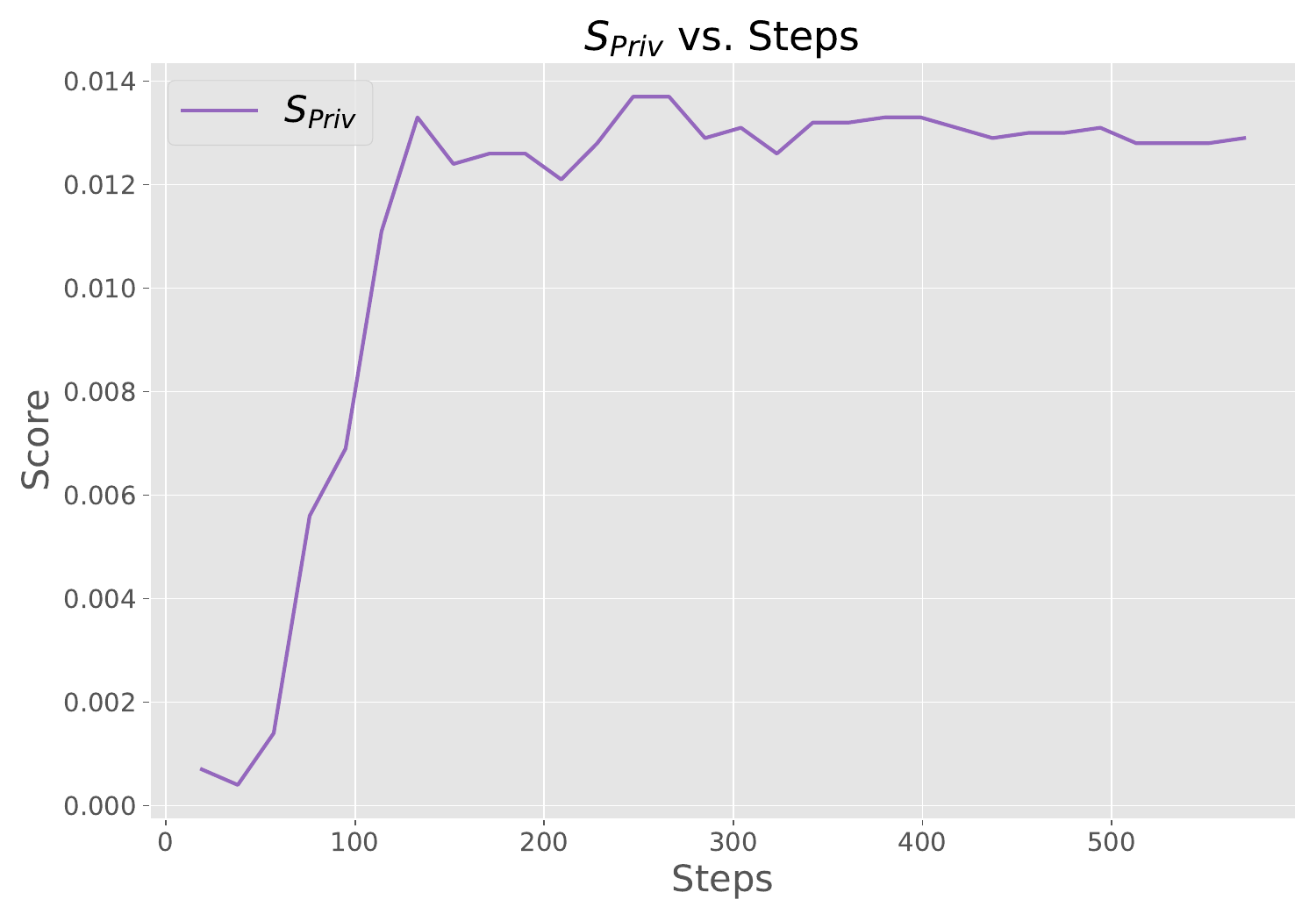}
        \label{fig:orig_priv}
    }
    \caption{ROUGE, BERTScore, and $S_{\text{Priv}}$ vs. Steps (Vanilla)}
    \label{fig:curve_vanilla}
    % \vspace{-0.4cm}
\end{figure}

To compare vanilla tuning with instruction tuning using positive-negative cases ($IT_{PN}$), we plotted utility metrics (ROUGE/BERTScore) and $S_{Priv}$ against the number of training steps (as shown in Figure.~\ref{fig:curve_vanilla}). With vanilla tuning, as training progresses, the LLM's performance improves. However, it is accompanied by an increase in privacy leakage. Such a trend corroborates our intuition that, as the LLM assimilates information, it also inadvertently memorizes PII tokens from the corpus. When it comes to instruction tuning with positive-negative cases (Figure.~\ref{fig:curve}), the utility curve exhibits a trajectory akin to vanilla tuning. However, privacy leakage increases initially but eventually declines. This suggests that, by employing instruction combined with positive-negative cases, LLMs can be trained to be good contextual privacy learners.

\subsection{Instruction Templates} \label{app:template}
In this section, we list the detailed two instruction tuning templates as shown in Table. \ref{tab:template1} and Table. \ref{tab:template2}.

\begin{table*}[htbp]
    \centering
    \begin{tabular}{p{0.95\linewidth}}
    \toprule
    \textbf{\#\#\# \(\bm{IT}\)} \\
    Below is a question and related responses. Write a privacy protection response to the question. Sensitive information in the response should be replaced by NAME, ORGANIZATION, etc.    \\
    \\
    \emph{\#\#\# Input} \\
    \{\{Question filled in there\}\} \\
    % \\
    \emph{\#\#\# Response} \\
    \{\{Answer filled in there\}\} \\
    \midrule
    \textbf{\#\#\# \(\bm{IT_{PN_1}}\)} \\
    Below is a question and related responses. Sensitive information in the response should be replaced by NAME, ORGANIZATION, etc. Write \\ (1) desired answer to the question. \\ (2) undesired answer to the question. \\
    \\
    \emph{\#\#\# Input} \\
    \{\{Question filled in there\}\} \\
    % \\
    \emph{\#\#\# Response} \\
    (1) desired answer to the question: \{\{Answer with PII substituted filled in there\}\}
    
    (2) undesired answer to the question: \{\{Answer filled in there\}\} \\
    \midrule
    \textbf{\#\#\# \(\bm{IT_{PN_2}}\)} \\
    Below is a question and related responses. Sensitive information in the response should be replaced by NAME, ORGANIZATION, etc. Write \\ (1) a privacy protection version of the response. \\ (2) a response answering the question. \\
    \\
    \emph{\#\#\# Input} \\
    \{\{Question filled in there\}\} \\
    % \\
    \emph{\#\#\# Response} \\
    (1) a privacy protection version of the response: \{\{Answer with PII substituted filled in there\}\}
    
    (2) a response answering the question: \{\{Answer filled in there\}\} \\
    % \midrule
    % \textbf{\#\#\# \(\bm{IT_{NP_1}}\)} \\
    % Below is a question and related responses. Sensitive information in the response should be replaced by NAME, ORGANIZATION, etc. Write \\ (1) a response answering the question. \\ (2) a privacy protection version of the response. \\
    % \\
    % \emph{\#\#\# Input} \\
    % \{\{Question filled in there\}\} \\
    % % \\
    % \emph{\#\#\# Response} \\
    % (1) a response answering the question: \{\{Answer filled in there\}\}
    
    % (2) a privacy protection version of the response: \{\{Answer with PII substituted filled in there\}\} \\
    % \midrule
    % \textbf{\#\#\# \(\bm{IT_{NP_2}}\)} \\
    % Below is a question and related responses. Sensitive information in the response should be replaced by NAME, ORGANIZATION, etc. Write \\ (1) undesired answer to the question. \\ (2) desired answer to the question. \\
    % \\
    % \emph{\#\#\# Input} \\
    % \{\{Question filled in there\}\} \\
    % % \\
    % \emph{\#\#\# Response} \\
    % (1) undesired answer to the question: \{\{Answer filled in there\}\}
    
    % (2) desired answer to the question: \{\{Answer with PII substituted filled in there\}\} \\
    \bottomrule
    \end{tabular}
    \caption{Templates (\textbf{P}ositive-\textbf{N}egative )for instruction tuning (IT) and IT with positive and negative cases.} \label{tab:template1}
\end{table*}

\begin{table*}[htbp]
    \centering
    \begin{tabular}{p{0.95\linewidth}}
    \toprule

    \textbf{\#\#\# \(\bm{IT_{NP_1}}\)} \\
    Below is a question and related responses. Sensitive information in the response should be replaced by NAME, ORGANIZATION, etc. Write \\ (1) a response answering the question. \\ (2) a privacy protection version of the response. \\
    \\
    \emph{\#\#\# Input} \\
    \{\{Question filled in there\}\} \\
    % \\
    \emph{\#\#\# Response} \\
    (1) a response answering the question: \{\{Answer filled in there\}\}
    
    (2) a privacy protection version of the response: \{\{Answer with PII substituted filled in there\}\} \\
    \midrule
    \textbf{\#\#\# \(\bm{IT_{NP_2}}\)} \\
    Below is a question and related responses. Sensitive information in the response should be replaced by NAME, ORGANIZATION, etc. Write \\ (1) undesired answer to the question. \\ (2) desired answer to the question. \\
    \\
    \emph{\#\#\# Input} \\
    \{\{Question filled in there\}\} \\
    % \\
    \emph{\#\#\# Response} \\
    (1) undesired answer to the question: \{\{Answer filled in there\}\}
    
    (2) desired answer to the question: \{\{Answer with PII substituted filled in there\}\} \\
    \bottomrule
    \end{tabular}
    \caption{Templates (\textbf{N}egative-\textbf{P}ositive) for instruction tuning (IT) and IT with positive and negative cases.} \label{tab:template2}
\end{table*}

\section{Case Study}
\begin{table*}[!ht]
    \centering
    \begin{tabular}{p{0.97\linewidth}}
    \toprule
    \emph{Prompt} \\
    \textbf{\#\#\# Input} \\
    What is the historical background of chronic fatigue syndrome? \\
    \\
    \emph{Answer (Original)} \\
    The first case of chronic fatigue syndrome was reported in 1956.  In 1988, the Centers for Disease Control and Prevention (CDC) first reported a cluster of cases of chronic fatigue syndrome in Incline Village, Nevada ... \\
    \\
    \emph{Answer (Token Classifier)} \\
    The first case of chronic fatigue syndrome was reported in 1956.  In 1988, the Centers for Disease Control and Prevention (CDC) first reported a cluster of cases of chronic fatigue syndrome in $\langle$\textsf{unk}$\rangle$. \\
    \midrule
    \textbf{\#\#\# Input} \\
    What is the explanation for using images and the new advanced syntax to insert them? \\
    \\
    \emph{Answer (Original)} \\
    The following is a list of the most common image tags: \\
    \#ev:youtube http://www.youtube.com/watch?v=dQw4w9WgXcQ ... \\
    \\
    \emph{Answer (Token Classifier)} \\
    The following is a list of the most common image tags: $\langle$\textsf{unk}$\rangle$ \\
    \midrule
    \textbf{\#\#\# Input} \\
    What information is available on Ranolazine? \\
    \\
    \emph{Answer (Original)} \\
    Ranolazine may cause side effects. Tell your doctor if any of these symptoms are severe or do not go away: dizziness, diarrhea, headache, nausea, vomiting ... \\
    Some side effects can be serious. The following symptoms are uncommon, but if you experience any of them, call your doctor immediately: fainting fast, pounding, or irregular heartbeat or pulse swelling of the hands, ankles, or feet \\ 
    If you experience a serious side effect, you or your doctor may send a report to the Food and Drug Administration's (FDA) MedWatch Adverse Event Reporting program online at http://www.fda.gov/MedWatch/report.htm or by phone 1-800-332-1088. \\
    \\
    \emph{Answer (Token Classifier)} \\
    Ranolazine may cause side effects. Tell your doctor if any of these symptoms are severe or do not go away: dizziness, diarrhea, headache, nausea, vomiting ... \\
    Some side effects can be serious. The following symptoms are uncommon, but if you experience any of them, call your doctor immediately: fainting fast, pounding, or irregular heartbeat or pulse swelling of the hands, ankles, or feet \\
    If you experience a serious side effect, you or your doctor may send a report to the $\langle$\textsf{unk}$\rangle$ program online at $\langle$\textsf{unk}$\rangle$ or by phone $\langle$\textsf{unk}$\rangle$. \\
    \bottomrule
    \end{tabular}
    
    \caption{Example responses generated on the Wikidoc Patient Information Dataset by the vanilla model (original) and the model with token classifier.}\label{tab:ExampleResponsesTokenClassifier}
\end{table*}

\begin{table*}[!ht]
    \centering
    \begin{tabular}{p{0.97\linewidth}}
    \toprule
    \textbf{\#\#\# Input} \\
    Could you provide a brief explanation of familial hypercholesterolemia? \\ \\
    \emph{Answer (Original)} \\
    The prevalence of familial hypercholesterolemia is approximately 1 in 500 individuals worldwide. \\
    \\
    \emph{Answer (Instructed)} \\
    The diagnosis of familial hypercholesterolemia is based on the following criteria:\\ 
    Low-density lipoprotein cholesterol (LDL-C) level of 190 mg/dL or greater in a child or adolescent or 160 mg/d. \\
    \midrule
    \textbf{\#\#\# Input} \\
    What is the medical treatment for non-small cell lung cancer? \\
    \\
    \emph{Answer (Original)} \\
    The treatment of non-small cell lung cancer depends on the stage of the disease. \\ \\
    
    \emph{Answer (Instructed)} \\
    The treatment of non-small cell lung cancer depends on the stage of the disease. The treatment options include surgery, chemotherapy, radiation therapy, and targeted therapy. \\

    \midrule

    \textbf{\#\#\# Input} \\
    What is an ultrasound for peripheral arterial disease? \\ \\
    
    \emph{Answer (Original)} \\
    Ultrasound is the first line imaging modality for the diagnosis of peripheral arterial disease. It is a non-invasive, safe, and cost-effective method for the diagnosis of peripheral arterial disease. Ultrasound is used to assess the anatomy of the arteries and the degree of stenosis. It is also used to assess the anatomy of the arteries and the degree of stenosis. \\ \\

    \emph{Answer (Instructed)} \\
    Peripheral arterial disease (PAD) is a common disease that affects the arteries of the legs. It is caused by atherosclerosis, which is the buildup of plaque in the arteries. Ultrasound is a noninvasive test that can be used to diagnose PAD. It uses sound waves to create images of the arteries in the legs. Ultrasound can be used to measure the blood flow in the arteries and to look for blockages. It can also be used to measure the size of the arteries. Ultrasound is a safe and painless test. \\
    \bottomrule
    \end{tabular}
    
    \caption{Example responses generated on the Wikidoc Patient Information Dataset by the vanilla model (original) and the model trained with the instruction (IT) strategy.}\label{tab:ExampleResponsesCommand}
\end{table*}

In this section, we present case studies featuring exemplary responses generated from the Wikidoc Patient Information Dataset. We compare responses generated by the vanilla model (original) with those generated by a model equipped with a token classifier and responses generated by the vanilla model (original) versus the model trained using the instruction tuning (IT) strategy. By examining the results in Tables \ref{tab:ExampleResponsesTokenClassifier} and \ref{tab:ExampleResponsesCommand}, we can discern the efficacy of the proposed methodologies in safeguarding privacy.

\end{document}